\documentclass[lettersize,journal]{IEEEtran}
\usepackage{amsmath,amsfonts}
\usepackage{algorithmic}
\usepackage{algorithm}
\usepackage{array}
\usepackage[caption=false,font=normalsize,labelfont=sf,textfont=sf]{subfig}
\usepackage{textcomp}
\usepackage{stfloats}
\usepackage{url}
\usepackage{verbatim}
\usepackage{graphicx}
\usepackage{cite}
\usepackage{dirtytalk}
\usepackage{booktabs}
\usepackage{multirow}
\usepackage{rotating}
\usepackage{hyperref}
\hypersetup{
	colorlinks=true,
	linkcolor=cyan,
	filecolor=blue,      
	urlcolor=red,
	citecolor=green,
}
\usepackage[T1]{fontenc}

\newcommand{\etal}{\textit{et al.}}
\newcommand{\ie}{\textit{i.e.}}

\begin{document}

\title{A Survey on Video Action Recognition in Sports: Datasets, Methods and Applications}

\author{\IEEEauthorblockN{Fei Wu\IEEEauthorrefmark{1}, Qingzhong Wang\IEEEauthorrefmark{1}\IEEEauthorrefmark{2}, Jiang Bian\thanks{* The three authors contribute equally to this work.}\IEEEauthorrefmark{1},~\emph{Member, IEEE}, Haoyi Xiong\thanks{$\dagger$ Corresponding authors.}\IEEEauthorrefmark{2}, \emph{Senior Member, IEEE}, Ning Ding, Feixiang Lu, Jun Cheng, and Dejing Dou, \emph{Senior Member, IEEE}
\thanks{Fei Wu and Ning Ding are with the Department of Physical Education, Peking University, Beijing, China. Email: wufei@pku.edu.cn.}
\thanks{Qingzhong Wang, Jiang Bian, Haoyi Xiong, Feixiang Lu, Jun Cheng, and Dejing Dou are with Baidu Inc., Beijing, China. Email: qingzwang@outlook.com, bianjiang03@baidu.com, xionghaoyi@baidu.com.}
}
\thanks{This work was support in part by National Key R\&D Program of China (No. 2021ZD0110303). }}



\maketitle

\begin{abstract}
To understand human behaviors, action recognition based on videos is a common approach. Compared with image-based action recognition, videos provide much more information. Reducing the ambiguity of actions and in the last decade, many works focused on datasets, novel models and learning approaches have improved video action recognition to a higher level. However, there are challenges and unsolved problems, in particular in sports analytics where data collection and labeling are more sophisticated, requiring sport professionals to annotate data. In addition, the actions could be extremely fast and it becomes difficult to recognize them. Moreover, in team sports like football and basketball, one action could involve multiple players, and to correctly recognize them, we need to analyse all players, which is relatively complicated. In this paper, we present a survey on video action recognition for sports analytics. We introduce more than ten types of sports, including team sports, such as football, basketball, volleyball, hockey and individual sports, such as figure skating, gymnastics, table tennis, tennis, diving and badminton. Then we compare numerous existing frameworks for sports analysis to present status quo of video action recognition in both team sports and individual sports. Finally, we discuss the challenges and unsolved problems in this area and to facilitate sports analytics, we develop a toolbox using PaddlePaddle~\footnote{The toolbox can be found at \url{https://github.com/PaddlePaddle/PaddleVideo}}, which supports football, basketball, table tennis and figure skating action recognition.
\end{abstract}


\section{Introduction}



\begin{figure}[t]
    \centering
    \includegraphics[width=0.8\linewidth]{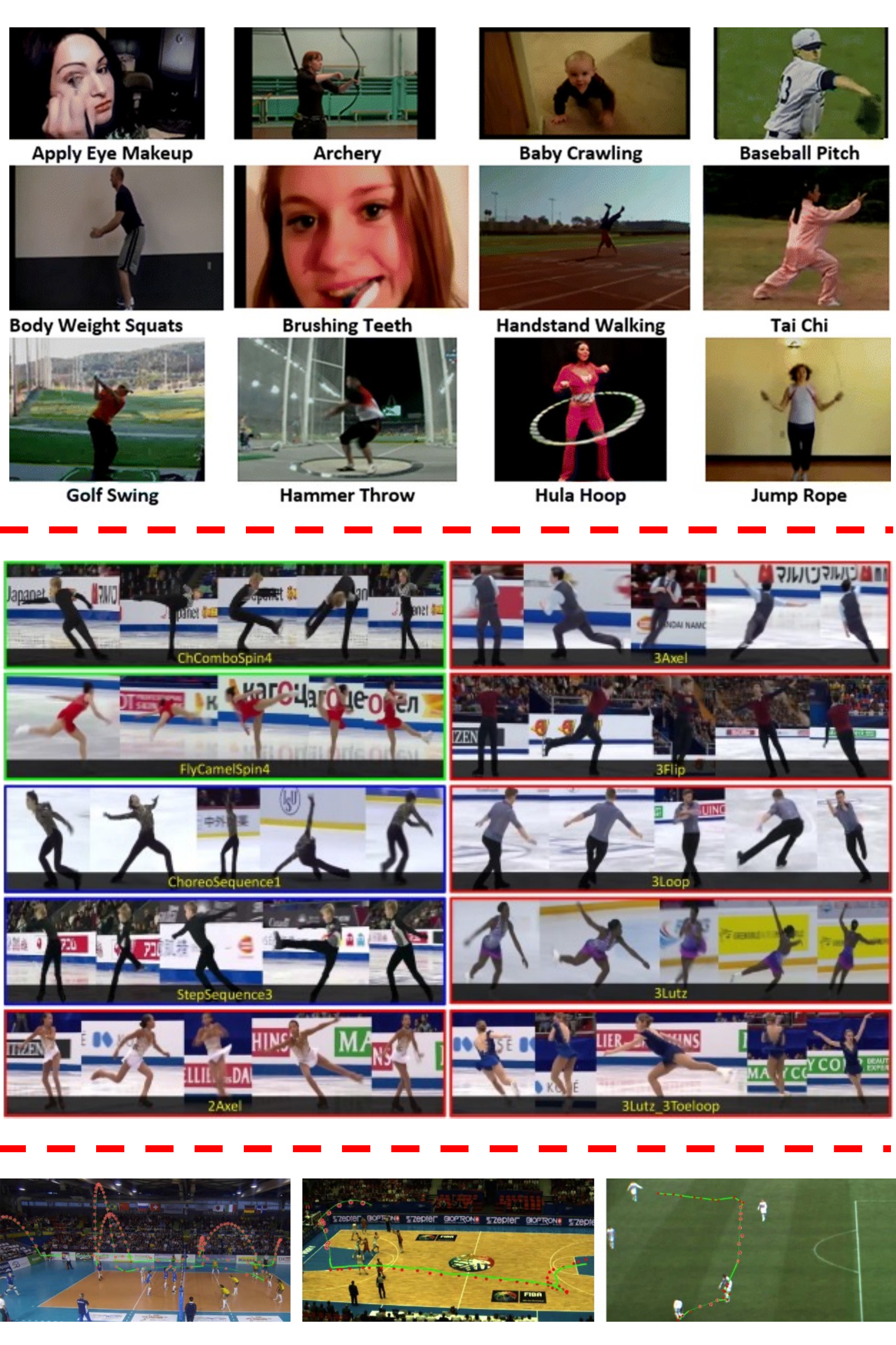}
    \caption{The comparison among common actions in our daily life, actions in individual sports and actions in team sports. Top: common actions in UCF101 \cite{soomro2012ucf101}, which is a coarse annotated dataset for action recognition. Middle: figure skating actions in FSD-10 \cite{liu2020fsd}, which is a fine-grained annotated figure skating dataset. Bottom: activities in volleyball, basketball and football \cite{maksai2016players}, where each action could involve multiple players.}
    \label{fig:action_comparison}
\end{figure}

\IEEEPARstart{T}he number of videos is rapidly increasing and there is a massive demand of analyzing them, namely video understanding, such as understanding the behaviors of people, tracking objects, recognizing abnormal behaviors, and content-based video retrieval. Thanks to the development of video understanding technologies, there are many applications in our everyday life, for example, surveillance systems. Action recognition lies at the heart of video understanding, which is an elementary module for analyzing videos. Researchers have put much effort on action recognition, labeling a large number of videos \cite{kuehne2011hmdb,soomro2012ucf101,caba2015activitynet,abu2016youtube,carreira2017quo,gu2018ava,goyal2017something} and proposing many impressive models to improve the recognition accuracy \cite{simonyan2014two,wang2018non,feichtenhofer2019slowfast,lin2019tsm,arnab2021vivit,li2020tea}. However, the popular datasets like ActivityNet \cite{caba2015activitynet} and Kinetics-400 \cite{kay2017kinetics} only consider the activities in our daily life, such as walking, driving cars and riding bikes. Although, some datasets contains sports-related activities, the labels are coarse and it is difficult to directly use them for specific sports analysis. In addition, to achieve the goal of fine-grained sports action recognition, we need to label videos that focus on specific sports, such as  football and basketball. Moreover, the fine-grained annotations normally require domain knowledge and professional players should be involved in video labeling. Figure~\ref{fig:action_comparison} shows the comparison between common actions in our daily life and actions in specific sports, such as figure skating and basketball. Obviously, to annotate an action as 3Axel or 3Flip, domain knowledge is required and professional players should be involved in data annotation, which is one challenge in sports video analysis. 

Recently, researchers in the communities of computer vision and sports pay much attention to sports video analysis, including building datasets and proposing novel methodologies \cite{giles2020machine,cust2019machine,hendry2020development,pickering2019development,russell2021moving,rangasamy2020deep,ibrahim2018hierarchical,giancola2018soccernet,martin2020fine,martin2021sports,cioppa2020context,li2021multisports,qi2019stagnet,mahaseni2021spotting,liu2020fsd}. In most existing works on sports video analysis, recognizing the actions of players in videos is crucial. On one hand, recognizing the group activities is able to assist coaches to make better decisions and players to understand their performances on fulfilling the coaches' strategies. On the other hand, recognizing the individual actions can benefit training players via correcting the small action errors \cite{bertasius2017baller,sri2021toward}. Another wide application of sports action recognition is in sports TV programs, where there is a massive demand of highlights generation and action recognition can significantly improves the localization accuracy \cite{tejero2018summarization,shukla2018automatic,zhao2019visual,yan2021new}. 

\begin{figure*}[t]
    \centering
    \includegraphics[width=\linewidth]{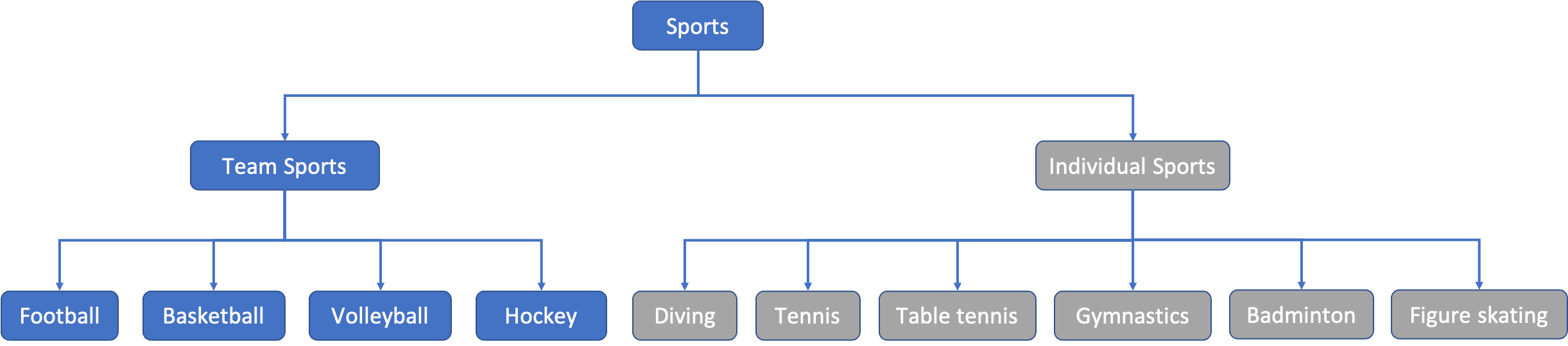}
    \caption{An example of sports categories.}
    \label{fig:sport_cat}
\end{figure*}

\begin{table}[t]
    \centering
    \begin{tabular}{c|c|c}
    \hline
         Sport& 2011-2016 &2017-present \\
         \hline
         Football &\cite{wu2013action,bialkowski2014large,wang2014take,durus2014ball,chen2014play,woods2016discriminating} &\cite{tsunoda2017football,ullah2017action,yu2018comprehensive,khan2018learning,ganesh2019novel,gerats2020individual,vanderplaetse2020improved,sanford2020group,mahaseni2021spotting,spitz2021video,koshkina2021contrastive} \\
         Basketball &\cite{chen2012recognizing,direkoglu2012team,sampaio2015exploring,bettadapura2016leveraging,bilen2016dynamic,ramanathan2016detecting,chauhan2016automatic} &\cite{acuna2017towards,wu2019ontology,arbues2019single,chen2020analysis,gu2020fine,wu2020fusing,vzemgulys2020recognition,ma2021npu,liu2021recognition,li2021research,lin2021lightweight,junjun2021basketball,fu2021camera}\\
         Volleyball &\cite{urgesi2012long,waltner2014indoor,vales2015saeta,amer2015sum,ibrahim2016hierarchical} & \cite{kautz2017activity,haider2019evaluation,salim2019volleyball,suda2019prediction,thilakarathne2021pose,tian2021optimization} \\
         Hockey &\cite{mukherjee2011recognizing,bermejo2011violence,wang2012discriminative,xu2014violent,routley2015markov,senst2015local} &\cite{fani2017hockey,mukherjee2017fight,sozykin2018multi,neher2018hockey,cai2019temporal,vats2019two,song2019novel,rangasamy2020hockey,vats2021puck,vats2021ice}\\
         Diving &\cite{napolitano2014cliff} & \cite{li2018resound,kanojia2019attentive,nekoui2020falcons,kumawat2021depthwise,zhi2021mgsampler,wang2021bevt,bertasius2021space} \\
         Tennis &\cite{farajidavar2011transductive,connaghan2011game,zhou2014tennis,vainstein2014modeling,reno2015tennis,reno2016real} &\cite{reno2017technology,vinyes2017deep,cai2018rgb,shimizu2019prediction,skublewska2019recognition,singanporia2019recognition,skublewska2020learning,cai2020deep,ullah2021attention,ning2021deep} \\
         Table tennis &\cite{wong2011tracking,tamaki2013reconstruction,draschkowitz2015using,heo2015analysis,myint2015tracking,myint2016tracking} &\cite{martin2018sport,martin2019optimal,hegazy2020ipingpong,martin20213d,kong2021ai,zahra2021two,aktas2021spatiotemporal,martin2021three,kulkarni2021table}\\
         Gymnastics &\cite{li2013real,potop2013learning,ylenia2013assessment,bouazizi2014effects,omorczyk2015high,khong2016simple} &\cite{shao2020finegym,hong2021video,duan2021revisiting,chen2021sportscap} \\
         Badminton & \cite{davar2011domain,teng2011detection,careelmont2013badminton,ting2014automatic,ramasinghe2014recognition,shan2015investigation} &\cite{ting2016potential,chu2017badminton,weeratunga2017application,ting2017badminton,ghosh2018towards,yunwei2019video,rahmad2019badminton,tao2020extracting,fang2021motion,yoshikawa2021shot}\\
         Figure Skating &\cite{haraguchi2011development,mazurkiewicz2015biomechanics} &\cite{xu2019learning,nakano2020estimating,liu2020fsd,tian20203d,tian2020multi,shan2020fineskating,li2021spatial,liu2021temporal,xia2022skating} \\
    \hline
    \end{tabular}
    \caption{A list of representative research works on sports video analysis in last decade.}
    \label{tab:ref}
\end{table}

However, there are many types of sports and each type of sports requires a specific model. Normally, we can roughly classify sports into team sports -- individuals are organized into opposing teams which compete to win and individual sports -- participants compete as individuals. In Figure~\ref{fig:sport_cat}, we present an example of sports categories. The analytics of team sports like football and individual sports such as diving are different. For team sports, each action could involve multiple players (see Figure~\ref{fig:action_comparison}) and each player has a specific action, such as dive and screen in basketball. In addition, the trajectory of the ball and the interaction between the ball and players are important in team sports analysis, hence, to accurate recognize the actions in team sports, we need to tracking the ball, multiple players and modeling the interactions \cite{maksai2016players}. While in individual sports, there is only one players such as gymnastics or two players such as tennis and badminton and to recognize the actions in individual sports, we can only focus on one or two players via person detection. In the last decade, there are many works on various types of sports and we present some representative works in table \ref{tab:ref}, where we can see that more works emerge in the recent five years and researchers start paying attention to the sports that receive less attention, such as diving and figure skating.

In this paper, we focus on video action recognition in various sports. One of the most related work is proposed by Y. Zhu \etal \cite{zhu2020comprehensive} -- a study of deep video action recognition, but it does not pay much attention to sports. D. Tan \etal  \cite{rahmad2018survey} review video-based action recognition approaches in badminton, such as recognizing the actions of service and smashing, while team sports and other individual sports are not considered and the popular datasets used for action recognition are not introduced. Although J. Gudmundsson \etal \cite{gudmundsson2017spatio}, R. Bonidia \etal \cite{bonidia2018computational} and R. Beal \etal \cite{beal2019artificial} review multiple sports, they pay much attention on sports data mining instead of video action recognition. M. Manafifard \etal \cite{manafifard2017survey} propose a survey on player tracking in soccer videos, which also reviews video technologies like object tracking and detection, however, only soccer is taken into account. H. Shih \cite{shih2017survey} proposes a survey on video technologies in content-aware sports analysis, such as object and video event detection, while we focus on action recognition in sports and provide a deep learning toolbox that supports figure skating, football, basketball and table tennis action recognition, which is publicly available.

To sum up, the contributions of the survey are in three folds.
\begin{itemize}
    \item First, we focus on the key part of sports video understanding -- action recognition and introduce more than ten sports, including team sports like football, basketball, volleyball, hockey and individual sports such as diving, tennis, gymnastics and table tennis.
    \item Second, we provide a sports genre classification and roads maps of action recognition methods in different types of sports. In addition, we present a summary of sports-related datasets for action recognition.
    \item Third, we present current state of video action recognition in different types of sports and the challenges should be paid attention to in the future. Moreover, to facilitate researches in sports video action recognition, we provide a deep learning toolbox that supports video action recognition in multiple sports, which is publicly available at \url{https://github.com/PaddlePaddle/PaddleVideo}.
\end{itemize}

The rest of the paper is organized as follows. In section \ref{sec:dataset}, we introduce the sports-related datasets used for action recognition. We present the survey of methodologies for individual action recognition in section \ref{sec:individual}, while in section \ref{sec:team}, we review the methodologies for team activity recognition. In section \ref{sec:application}, we summarize the applications of video action recognition in sports, such as education and coaching. Section \ref{sec:challenge} summarizes the challenges that should be paid more attention in the future. Last but not least, we make conclusions in section \ref{sec:conclusion}.

\section{Sports-related Datasets}\label{sec:dataset}

\begin{table*}[!htp]
    \centering
    \begin{tabular}{c|c|c|c|c|c|c|c}
    \hline
        Dataset & Sport & Year &Task & \# Videos & Avg. length & \# Categories & Publicly Available \\
    \hline
        CVBASE Handball \cite{pers2005cvbase} &handball & 2006 & cls. & 3 & 10m &- & Yes \\ \hline
        CVBASE Squash \cite{pers2005cvbase} &squash & 2006 & cls. & 2 & 10m &- & Yes \\ \hline
        UCF sports \cite{rodriguez2008action} & multiple & 2008 & cls. &150 &6.39s & 10 & Yes \\ \hline
        APIDIS \cite{de2008distributed,parisot2019consensus} & basketball & 2008 & det.\& cls. &- &- &- & Yes \\ \hline
        Soccer-ISSIA \cite{d2009semi} & football & 2009 & tra. & - & - & - & Yes \\ \hline
        MSR Action3D \cite{li2010action} &multiple &2010 &cls. &567 &- &20 & Yes \\ \hline
        Olympic \cite{niebles2010modeling} & multiple &2010 & cls. & 800 &- &16 &Yes \\ \hline
        Hockey Fight \cite{bermejo2011violence} & hockey & 2011 & cls. & 1,000 &- & 2 & Yes \\ \hline
        ACASVA \cite{de2011evaluation} & tennis & 2011 & cls. & 6 & - & 4 &Yes \\ \hline
        THETIS \cite{gourgari2013thetis} & tennis & 2013 & cls. &1,980 &- & 12 & Yes \\ \hline
        Sports 1M \cite{karpathy2014large} & multiple &2014 &cls. & 1M & 36s &487 & Yes\\ \hline
        OlympicSports \cite{pirsiavash2014assessing} &multiple &2014 & ass. &309 &- & 2 & Yes\\ \hline
        SVW \cite{safdarnejad2015sports} &multiple &2015 &det.\& cls. & 4,100 &11.6s & 44 &Yes \\ \hline
        Basket-1,2 \cite{maksai2016players} & basketball &2016 & det.\& cls. & - & - &4 & No \\ \hline
        Volleyball-1,2 \cite{maksai2016players} & volleyball & 2016 & det.\& cls. &-&-&- &No \\ \hline
        HierVolleyball \cite{ibrahim2016hierarchical} & volleyball & 2016 & det.\& cls. &- &- &- & Yes \\ \hline
        HierVolleyball-v2 \cite{DBLP:journals/corr/IbrahimMDVM16} & volleyball & 2016 & det.\& cls. &- &- &- & Yes \\ \hline
        NCAA \cite{ramanathan2016detecting} &basketball &2016 & cls.\& loc. &14,548 & 4s &11 &Yes \\ \hline
        Football Action \cite{tsunoda2017football} & football &2017 & cls. &3,281 &- & 5 & No \\ \hline
        TenniSet \cite{faulkner2017tenniset} & tennis & 2017 & loc.\& cls. & 5 &- &6 & Yes \\ \hline
        OlympicScoring \cite{parmar2017learning} &multiple &2017 & ass. & 716 & - & 3 & Yes\\ \hline
        Soccer Player \cite{lu2017light} & football &2017 & tra.\& det. &- &- &- & Yes \\ \hline
        SPIROUDOME \cite{parisot2017scene} &basketball &2017 & det. &- &- &- &Yes \\ \hline
        SpaceJam \cite{francia2018classificazione} & basketball &2018 &cls. &15 &1.5h &10 & Yes\\ \hline
        Diving48 \cite{li2018resound} &diving & 2018 & cls. &18,404 & - & 48 & Yes \\ \hline
        ComprehensiveSoccer \cite{yu2018comprehensive} & football & 2018 & det.\& cls. & 220 & 0.77h &- &Yes \\ \hline
        TTStroke-21 \cite{martin2018sport} & table tennis &2018 & cls. &129 &43m &21 &Yes \\ \hline
        SoccerNet \cite{giancola2018soccernet} &football & 2018 & loc.\& cls. & 500 &1.5h &3 & Yes\\ \hline
        Badminton Olympic \cite{ghosh2018towards} & badminton &2018 &loc.\& cls. &10 & 1h & 12 & Yes \\ \hline
        SPIN \cite{schwarcz2019spin} & table tennis &2019 & tra.\& cls. &- &- &- & No\\ \hline
        GolfDB \cite{mcnally2019golfdb} & golf &2019 & cls. & 1,400 &- &8 &Yes \\ \hline
        AQA \cite{parmar2019action} & multiple & 2019 & ass. &1,189 &- &7 &Yes \\ \hline
        MTL-AQA \cite{parmar2019and} &diving &2019 & ass. &1,412 &- &- &Yes \\ \hline
        OpenTTGames \cite{voeikov2020ttnet} & table tennis & 2020 &seg.\& det. & 12 &- &- & Yes \\ \hline
        FineGym \cite{shao2020finegym} & gymnastics & 2020 & cls.\& loc. & - &- &288 & Yes\\ \hline
        SSET \cite{feng2020sset} & football & 2020 &tra.\& det. &350 &0.8h & 30 & Yes\\ \hline
        SoccerDB \cite{jiang2020soccerdb} & football & 2020 &cls.\& loc. &346 &1.5h &11 &Yes \\ \hline
        FineBasketball \cite{gu2020fine} & basketball &2020 & cls. &3,399, &- & 26 &Yes \\ \hline
        FSD-10 \cite{liu2020fsd} & figure skating & 2020 & ass.\& cls. &- &- &10 &Yes \\ \hline
        FineSkating \cite{shan2020fineskating} & figure skating & 2020 & ass.\& cls. & 46 & 1h &- &Yes \\ \hline
        MCFS \cite{liu2021temporal} & figure skating &2021 &loc.\& cls. & 11,656 & - & 130 & Yes\\ \hline
        Stroke Recognition \cite{kulkarni2021table} & table tennis & 2021 & cls. & 22,111 & - & 11 & Yes\\ \hline
        MultiSports \cite{li2021multisports} &multiple & 2021 &loc.\& cls. &3,200 &20.9s & 66 &Yes \\ \hline
        Player Tracklet \cite{vats2021player} & hockey &2021 &tra. &84 &36s &- & Yes \\ \hline
        NPUBasketball \cite{ma2021npu} & basketball &2021 &cls. & 2,169 & - & 12 & Yes \\ \hline
        SoccerNet-v2 \cite{deliege2021soccernet} & football & 2021 &loc.\& cls. & 500 &1.5h & 17 & Yes\\ \hline
        Win-Fail \cite{parmar2022win} &multiple &2022 & cls. & 1,634 & 3.3 &2 &Yes \\ \hline
        Stroke Forecasting \cite{wang2021shuttlenet} &badminton & 2022 & cls. & 43,191 & - & 10 & Yes \\ \hline
        FenceNet \cite{zhu2022fencenet} & fencing &2022 & cls. &652 & - & 6 &Yes \\
    \hline
    \end{tabular}
    \caption{A list of sports-related datasets used in the published papers. Note that some of them are not publicly available and ``multiple'' means that the dataset contains various sports instead of only one specific type of sports. ``det.'', ``cls.'', ``tra.'', ``ass.'', ``seg.'', ``loc.'' stand for player/ball detection, action classification, player/ball tracking, action quality assessment, object segmentation and temporal action localization, respectively. More details of the dataset can be found in section \ref{sec:dataset}.}
    \label{tab:dataset}
\end{table*}

Datasets are required to facilitate model training and evaluation, in particular in the era of deep learning since deep models are normally data-hungry. Researchers have put much effort into developing new sports-related datasets. Generally, to construct a dataset for sports video action recognition, we need to (1) define the type of sports that we want to investigate and the categories of actions in the specific sport, (2) collect videos from multiple sources, such as the internet and self-recorded videos, (3) process the collected videos like trimming and then annotate the processed videos. The annotations could vary based on the goal of the proposed dataset, but it should provide trimmed videos and the corresponding labels or untrimmed videos with the start and end time of each action and the action category. In some datasets, the annotation process could be more complicated. For example, apart from annotating action labels and temporal positions, bounding boxes of objects that impose the actions are also annotated in AVA dataset \cite{li2020ava}. In this section, we provide a comprehensive review of sports-related datasets and the list of datasets is shown in table \ref{tab:dataset}.

\subsection{Football}
Football is one of the most popular sports in the world and researchers pay much attention to football activity recognition, developing numerous datasets with different scales. 

\textbf{Soccer-ISSIA} \cite{d2009semi} is a relatively small dataset, composed of 18,000 high resolution frames recorded by 6 static cameras. The recorded videos are first automatically processed to extract blobs that indicate moving players and then the annotated bounding boxes are validated by humans. \textbf{Soccer-ISSIA} \cite{d2009semi} are normally used for player tracking, detection and team activity recognition. Similarly, \textbf{Soccer Player} \cite{lu2017light} is developed for player detection and tracking, comprising of 2,019 annotated frames with 22,586 player bounding boxes.

\textbf{Football Action} \cite{tsunoda2017football} is a private dataset composed of self-recorded videos that are captured using 14 synchronized and calibrated Full HD cameras and the position of each player is annotated using a bounding box. There are five categories of activities: pass, shoot, loose clearance and dribble.

\textbf{ComprehensiveSoccer} \cite{yu2018comprehensive} is composed of 222 broadcast videos and 170 video hours in total. The dataset is annotated in 3 level: positions of players using bounding boxes, event and story annotation at a coarse granularity and temporal annotations of shots. Totally, there are 11 categories of event, 15 types of story and 5 types of shot. The dataset can be used for various tasks in football video analysis, such as action classification, localization and player detection.

\textbf{SoccerNet} \cite{giancola2018soccernet} is a large-scale dataset for football action recognition and localization. There 500 complete soccer match videos collected from European leagues during 2014-2017. The total number of temporal annotations is 6,637 and the label of each temporal annotation is one of three categories: goal, substitution and yellow or red card. The actions are relatively sparse in \textbf{SoccerNet}, \ie, there are only 8.7 actions per hour on average.

\textbf{SSET} \cite{feng2020sset} is three times smaller than \textbf{SoccerNet}, comprising of 350 football match videos, totaling 282 video hours. Similar to \textbf{ComprehensiveSoccer}, the annotations are in three levels: bounding boxes of players, event/story categories and shot categories, but \textbf{SSET} is larger than \textbf{ComprehensiveSoccer} dataset.

\textbf{SoccerDB} \cite{jiang2020soccerdb} is in the same scale as \textbf{SoccerNet}, which is composed of 171,191 video segments trimmed from 346 soccer match videos and the total length of the videos is 668.6 hours. \textbf{SoccerDB} also annotates the positions of players using bounding boxes, which contains 702.096 bounding boxes. 11 labels are taken into account for activity annotation, including goal, foul, injured, red/yellow card, shot, substitution, free kick, corner kick, saves, penalty kick and background. Each segment belongs to one category and has a time boundary. In addition, 17,115 highlights in soccer match videos are also annotated, therefore, the dataset can be used for player detection, activity recognition, activity localization and highlight detection.

\textbf{SoccerNet-v2} \cite{deliege2021soccernet} extends \textbf{SoccerNet} \cite{giancola2018soccernet} via re-labeling the 500 untrimmed videos. In \textbf{SoccerNet}, there are only 3 categories, while \textbf{SoccerNet-v2} has 17 categories, such as throw in, foul, indirect free kick, corner, shots on target, shots off target, direct free kick, clearance, substitution, kick off, offside, yellow card, red card, goal, penalty, yellow-to-red card and ball out of play. Moreover, the actions in  \textbf{SoccerNet-v2} are much denser than these in \textbf{SoccerNet}, for example, there is one action every 25 seconds in \textbf{SoccerNet-v2}, whereas, there is only 8.7 actions per hour in \textbf{SoccerNet}. Similar to \textbf{SoccerNet}, \textbf{SoccerNet-v2} can be employed for action recognition and localization.

Basically, large-scale datasets+deep models dominate the field of soccer video action recognition in recent years, increasing the popularity of \textbf{SoccerNet} \cite{giancola2018soccernet} and \textbf{SoccerNet-v2} \cite{deliege2021soccernet}. While \textbf{SoccerDB} \cite{jiang2020soccerdb}, \textbf{SSET} \cite{feng2020sset} and \textbf{ComprehensiveSoccer} \cite{yu2018comprehensive} are more feasible for the tasks that require player detection.

\subsection{Basketball}
Basketball has drawn much attention from researches owing to its popularity in the world and numerous basketball datasets at different scales have been developed.

\textbf{APIDIS} \cite{de2008distributed,parisot2019consensus} is composed of seven videos of the same basketball match, which is recorded by seven calibrated cameras located in different positions of the basketball court. The positions of players and ball are annotated using bounding boxes. Clock and non-clock actions are also annotated, such as throw, violation, foul, pass, positioning and rebound. Each action has a time boundary and a label, thus, \textbf{APIDIS} can be used for both player detection and basketball action recognition. The dataset is challenging since the contrast between the background and players is low \cite{lu2017light}.

\textbf{Basket-1,2} \cite{maksai2016players} contains two basketball frame sequences -- one has 4000 frames captured by 6 cameras and another has 3000 frames captured by 7 cameras. The cameras are synchronized and each can capture 25 frames per second. There are four action categories in the dataset: possessed ball, passed ball, flying ball, and ball out of play. \textbf{Basket-1,2} can be used for basketball action recognition and ball detection.

\textbf{NCAA} \cite{ramanathan2016detecting} is a relatively large dataset for basketball action recognition, composed of 257 untrimmed NCAA game videos and the video length are normally in 1.5 hours. After processing, the dataset comprises 14,548 video segments with time boundary, each of which contains a action belongs to one of 14 categories, such as 3-pint success, 3-point fail, steal, slam dunk success and slam dunk fail. In addition, \textbf{NCAA} also provides 9,000 frames with bounding boxes of players, therefore, people can also use it for player detection.

\textbf{SPIROUDOME} \cite{parisot2017scene} is similar to \textbf{APIDIS}, where the videos are captured using 8 cameras. The positions of players are annotated using bounding boxes, therefore, \textbf{SPIROUDOME} is generally employed for player detection.

\textbf{SpaceJam} \cite{francia2018classificazione} comprises 10 categories of basketball actions, including step, race, block, dribble, ball in hand, shooting, position, walk, defensive position and no action. \textbf{SpaceJam} collects 15 videos of the NBA championship and the Italian championship from YouTube and the length of each video is 1.5 hours. Besides RGB images, the estimated poses of players are also provided. Normally, \textbf{SpaceJam} can be used to develop skeleton-based action recognition models.

\textbf{FineBasketball} \cite{gu2020fine} is developed for fine-grained basketball action recognition, containing three broad categories -- dribbling, passing and shooting, and 26 fine-grained categories, such as behind-the-back dribbling, cross-over dribbling, hand-off, one-handed side passing, lay up shot, one-handed dunk and block shot. There are 3,399 video segments in total and each category contains roughly 130 video segments on average. \textbf{FineBasketball} is challenging since the dataset is imbalanced, for example, there are 717 video segments belonging to crossover dribbling, while the class of follow-up shot only contains 12 video segments.

\textbf{NPUBasketball} \cite{ma2021npu} is composed of 2,169 self-recorded video clips of basketball actions performed by professional players and each video belongs to one of 12 categories: standing dribble, front dribble, moving dribble, cross-leg dribble, behind-the-back dribble, turning around, squat, run with ball, overhead pass (catch or shoot), one-hand shoot, chest pass (catch or shoot), and side throw. Different from \textbf{FineBasketball} and \textbf{SpaceJam}, \textbf{NPUBasketball} provides not only RGB frames, but also depth maps and skeleton of players, thus, it can be used for developing various types of action recognition models.

\subsection{Volleyball}

Though volleyball is a relatively popular sport in the world, there are only a few volleyball datasets and most of them are on small scales.

\textbf{Volleyball-1,2} \cite{maksai2016players} contains two sequences -- one comprises 10,000 frames and another is composed of 19,500 frames. The positions of ball is manually annotated using bounding boxes, however, detecting the ball is challenging since it moves fast and blurred after striking.

\textbf{HierVolleyball} \cite{ibrahim2016hierarchical} is developed for team activity recognition, containing 1,525 annotated frames from 15 YouTube volleyball videos. Each player has a action label defined as waiting, setting, digging, falling, spiking, blocking and others, and some players perform a group activity, such as set, spike and pass. 

\textbf{HierVolleyball-v2} \cite{DBLP:journals/corr/IbrahimMDVM16} extends \textbf{HierVolleyball}, comprising 4,830 annotated frames from 55 YouTube volleyball videos. There are 9 categories of players' actions: waiting, setting, digging, failing, spiking, blocking, jumping, moving and standing, and winpoint is also considered as a team activity category. The positions of players are also annotated using bounding boxes, and it can be used for both player detection and action recognition.

\subsection{Hockey}
\textbf{Hockey Fight} \cite{bermejo2011violence} is a proposed for binary classification: fight and non-fight in hockey games, composed of 1,000 video clips from National Hockey League (NHL) games. Each clip contains 50 frames and has a label indicates fight or non-fight.

\textbf{Player Tracklet} \cite{vats2021player} comprises 84 video clips from broadcast NHL games and the average length of the videos is 36s. The positions of players and referee in each frame are annotated with bounding boxes and identity labels like players' names and numbers. \textbf{Player Tracklet} can be applied for player tracking and identification.

\subsection{Tennis}
Tennis is an individual sport, attracting tens of millions of people and researchers have constructed various dataset for tennis video analysis.

\textbf{ACASVA} \cite{de2011evaluation} is develop for tennis action recognition, in particular for evaluating primitive players' action in tennis games, where there are six broadcast videos of tennis games and three categories of actions: hit, non-hit and serve. The positions of players and time boundaries of actions are labeled, however, the dataset only provides the extracted features of video clips instead of the original videos.

\textbf{THETIS} \cite{gourgari2013thetis} is composed of 1,980 self-recorded videos belongs to 12 tennis actions: four backhand shots (backhand, backhand with two hands, backhand slice, backhand volley), four forehand shots (forehand flat, forehand slice, forehand volley, forehand open stands), three service shots (service flat, service kick, service slice) and smash. Besides RGB frames, \textbf{THETIS} also provides 1,980 depth videos, 1,217 2D skeleton videos and 1,217 3D skeleton videos, so it can be used for developing multiple types of action recognition models.

\textbf{TenniSet} \cite{faulkner2017tenniset} comprises five tennis videos of 2012 London Olympic matches from YouTube and six categories of events are considered, such as set, hit and serve. The time boundary of each event is labeled, therefore, it can be used for both recognition and localization. Interestingly, \textbf{TenniSet} also provides textural descriptions of actions, such as ``quick serve is an ace'', so it can also be used for action retrieval.

\subsection{Table Tennis}
Similar to tennis, strokes in table tennis are important and multiple datasets have been developed for table tennis stroke recognition.

\textbf{TTStroke-21} \cite{martin2018sport} is composed of 129 self-recorded videos of 94-hour games in the egocentric perspective. There are 1,378 annotated actions, each of which belongs to one of 21 categories, such as serve backhand spin, forehand push, backhand block and forehand loop. Though the strokes in table tennis games are relatively fast, \textbf{TTStroke-21} is not a challenging dataset and one possible reason is that the videos have a high freme rate (120 FPS).

\textbf{SPIN} \cite{schwarcz2019spin} also comprises self-recorded videos captured by two high-speed cameras (150 FPS), totaling 53 hours and 7.5 million high-resolution (1024$\times$1280) frames. The positions of ball are annotated using bounding boxes and 30 locations of players' joints are also labeled using heatmaps (15 joints for each player) in each frame. The dataset can be used for multiple tasks like ball tracking, pose estimation and spin prediction based on the trajectory of ball and player's poses.

\textbf{OpenTTGames} \cite{voeikov2020ttnet} consists of 12 HD videos of table tennis games (5 videos for training and 7 short videos for testing). Ball coordinates are annotated in each frame and 4,271 events are labeled, each of which has a label -- ball bounces, net hits or empty events. In addition, 4 frames before each event and 12 frames after are labeled using segmentation masks, including human, table and scoreboard, hence, \textbf{OpenTTGames} can be used for semantic segmentation, ball tracking and event classification.

\textbf{Stroke Recognition} \cite{kulkarni2021table} is similar to \textbf{TTStroke-21} but much larger, composed of 22,111 trimmed videos and each video contains a stroke belongs to one of 11 categories. The dataset is less challenging, for example, random forest with 21 trees achieves the accuracy of 96.20\% \cite{kulkarni2021table}.

\textbf{P$^2$A} \cite{p2a2022} is one of the largest dataset for table tennis analysis, composed of 2,721 untrimmed broadcasting videos, and the total length is 272 hours. The authors annotate each stroke in videos, including the category of the stroke and the indices of the starting and end frames. Plus, the stroke labels are confirmed by professional players, including Olympic table tennis players. 

\subsection{Gymnastics}
There are few datasets for gymnastics and one recent work named \textbf{FineGym} \cite{shao2020finegym} is developed for gymnastic action recognition and localization, consisting of 303 videos with around 708-hour length. \textbf{FineGym} is annotated in a hierarchical manner, for example, there are four high-level event labels, 15 categories of action sets for 4 events and 530 categories of element actions. The time boundaries of actions and sub-actions are labeled, therefore, \textbf{Gymnastics} can be used for fine-grained action recognition and localization. The task of event/set-level action recognition and localization are relatively easy, while element-level action recognition and localization are much more challenging.

\subsection{Badminton}
\textbf{Badminton Olympic} \cite{ghosh2018towards} is composed of 10 videos of ``singles'' badminton matches from YouTube and each video is generally in one hour. There are multiple types of annotations in the dataset. First, the positions of players in 1,500 frames are annotated using bounding boxes. Second, 751 temporal locations of when a player wins a point are annotated. Third, the time boundaries and labels of strokes are annotated, where there are 12 categories of strokes, such as serve and lob. With three types of annotations, \textbf{Badminton Olympic} can be used for multiple tasks -- player detection, point localization, action recognition and localization.

\textbf{Stroke Forecasting} \cite{wang2021shuttlenet}  is a most recent dataset, consisting of 43,191 trimmed video clips and each video clip has a stroke belongs to one of 10 categories -- smash, push, clear, defensive shot, net shot, drive, drop, lob, long service and short service. In addition to badminton action recognition, the dataset can also be used for stroke forecasting, \ie, given previous stokes in a rally, the model should predict what the next stroke is.

\subsection{Figure skating}
There are three dataset proposed for figure skating action recognition in recent years -- \textbf{FSD-10} \cite{liu2020fsd}, \textbf{FineSkating} \cite{shan2020fineskating} and \textbf{MCFS} \cite{liu2021temporal}.

\textbf{FSD-10} \cite{liu2020fsd} comprises ten categories of figure skating actions (Change Combination Spin 4, Fly Camel Spin 4, Choreo Sequence 1, Step Sequence 3, Double Axel, Triple Axel, Triple Flip, Triple Loop, Triple Lutz, Triple Lutz-Triple Toeloop) and each action has 91-233 video clips, ranging from 3s to 30s. In addition to action labels, \textbf{FSD-10} also provides scores of actions for action quality assessment.

\textbf{FineSkating} \cite{shan2020fineskating} is composed of 46 videos of figure skating competitions in 2018 and 2019, each of which is around 1 hour long. The labels are designed in a hierarchical manner, \ie, event labels and action labels. There are seven event labels, such as jump and spin, and each event has multiple actions, for example, the event of jump contains 7 actions: Axel, Flip, Toeloop, Loop, Lutz, Salchow and Euler. Moreover, the start time, end time and score of each action are also labeled, hence, it can be used for both action recognition and action quality assessment.

\textbf{MCFS} \cite{liu2021temporal} consists of 11,656 video segments from 38 figure skating competitions, totaling 17.3 hours and 1.7 million frames. Similar to \textbf{FineGym} \cite{shao2020finegym}, \textbf{MCFS} has three-level annotations: 4 set (jump, spin, sequence, none), 22 subsets (Camel spin, Axel,$\cdots$) and 130 element actions (double Axel, double Flip, triple Axel,  $\cdots$). The time boundaries of actions are also annotated, so \textbf{MCFS} can be applied for action recognition and localization.

\subsection{Diving}
\textbf{Diving48} \cite{li2018resound} contains 16,067 diving video segments for training and 2,337 for testing, totaling 18,404 video segments and covering 48 fine-grained categories of diving. Each class of action is composed of multiple elements, such as backward take-off and half twist. Compared with existing datasets for action recognition, \textbf{Diving48} has a relatively low bias, which is more fair for model evaluation.

By contrast, \textbf{MTL-AQA} \cite{parmar2019and} is developed for diving action quality assessment, consisting of 1,412 samples and each sample is annotated with an action quality score, action class and textural commentary, therefore it can be used for multiple tasks, including action quality assessment and recognition.

\subsection{Multiple Types of Sports} 
There are several datasets supporting multiple sports classification, where each video has a label indicates the category of sports, such as football, basketball and gymnastics, and a model is supposed to classify the videos. Generally, these datasets are used for coarse classification.

\textbf{UCF sports} \cite{rodriguez2008action} is proposed in 2008, composed of 150 video clips with 10FPS. The length of videos ranges from 2.02s to 14.40s and there are 10 categories, including diving, golf swing, kicking, lifting, riding horse, running, skate boarding, swing bench, swing side and walking.

Two years later, W. Li \etal \cite{li2010action} develop \textbf{MSR Action3D}, which contains 576 sequences of depth maps instead of RGB frames and people can use it to recognize sports actions, such as tennis serve, tennis swing and golf swing. The videos are in \textbf{MSR Action3D} are self-recorded.

\textbf{Olympic} \cite{niebles2010modeling} is a relatively large dataset, including 800 videos for 16 categories like long jump, high jump, tennis serve, diving and vault, and each category has 50 videos. The videos in Olympic are from Youtube instead of self-recorded, therefore, occlusions and camera movements are involved in videos, being more challenging.

\textbf{Sports 1M} \cite{karpathy2014large} is a much larger dataset, containing around one million videos that are from YouTube and 487 categories. There are 1,000-3,000 videos fro each category, so that the distribution of videos is relatively balance. Moreover, the labels are designed in a hierarchical manner, \ie, the high-level nodes like team sports, ball sports, winter sports are used for coarse classification and the leaf nodes, such as eight-ball, nine-ball and blackball of billiards can be used for fine-grained classification. To some extent, using this million-scale dataset, we can alleviate the problem of data hungry in deep learning.

\textbf{SVW} \cite{safdarnejad2015sports} is a dataset for both action classification and detection, composed of 4,100 videos and 44 action categories belong to 30 types of sports, such as soccer, swimming, tennis and volleyball. One property of this dataset is that the videos are captured by smartphones from the view of coaches and the quality of the videos is normally lower than the broadcasting videos, resulting in challenges for action recognition.

Recently, \textbf{MultiSports} \cite{li2021multisports} is proposed for multi-person sports, which is more challenging since each activity can involve multiple players who can perform different actions. The dataset covers four team sports -- aerobic gymnastics, football, basketball and volleyball, and 66 categories of actions. There are 3,200 videos and 37,701 action instances. Apart from annotating video segments (temporal labels), \textbf{MultiSports} also provides bounding boxes of players involved in the activities, therefore, it can be used for action recognition, temporal and spacial localization.

Besides recognizing the actions in sports, some other datasets are proposed for action assessment, \ie, a model should not only recognize the actions, but also provide a score that indicates the quality of the action. \textbf{OlympicSports} \cite{pirsiavash2014assessing} is proposed to evaluate the quality of diving and figure skating actions, comprising of 159 diving videos and 150 figure skating videos from Youtube, while \textbf{OlympicScoring} \cite{parmar2017learning} extends it by collecting more videos and introducing more types of sports, which is composed of 370 diving videos, 170 figure skating videos and 176 vault videos. However, the number of  videos in \textbf{OlympicScoring} is still limited for deep learning based methods. In contrast, \textbf{AQA} \cite{parmar2019action} dataset includes seven categories of sports: synchronous diving--10m platform, singles diving--10m platform, synchronous diving--3m spring board, gymnastic vault, skiing, snowboarding and trampoline. There 1,189 videos in total.

Interestingly, \textbf{Win-Fail} \cite{parmar2022win} is proposed for recognizing win or fail of actions. Though actions could be very complex, the results of actions, \ie, win/fail can be recognized via reasoning on the movements of objects. \textbf{Win-Fail} is composed of 817 win-fail video pairs collected from multiple domains like trick-shots and internet win-fails.

\subsection{Others}
\textbf{CVBASE Handball} \cite{pers2005cvbase} is developed for handball action recognition, comprising three synchronized videos and each video is 10-minus long. The trajectories of seven players, team activities like offensive, defensive and individual actions like pass, shot are annotated. Similar to \textbf{CVBASE Handball}, \textbf{CVBASE Squash} \cite{pers2005cvbase} composed of two 10-minus videos of different matches also provides trajectories of players and categories of strokes, such as lob, drop and cross.

\textbf{GolfDB} \cite{mcnally2019golfdb} is proposed to facilitate the analysis of golf swings, consisting of 1,400 high-quality golf swing video segments belong to eight swing categories, such as toe-up, top, impact and so on. In addition to action labels, \textbf{GolfDB} also provides bounding boxes of players, player name and sex.

\textbf{FenceNet} \cite{zhu2022fencenet} is composed of 652 videos belong to 6 categories -- rapid lunge, incremental speed lunge, with waiting lunge, jumping sliding lunge, step forward, and step backward. The actions are performed by expert-level fencers. In addition to RGB frames, the dataset also provides 3D skeleton data and depth data.

\section{Individual Action Recognition}\label{sec:individual}

\begin{figure}[t]
    \centering
    \includegraphics[width=0.8\linewidth]{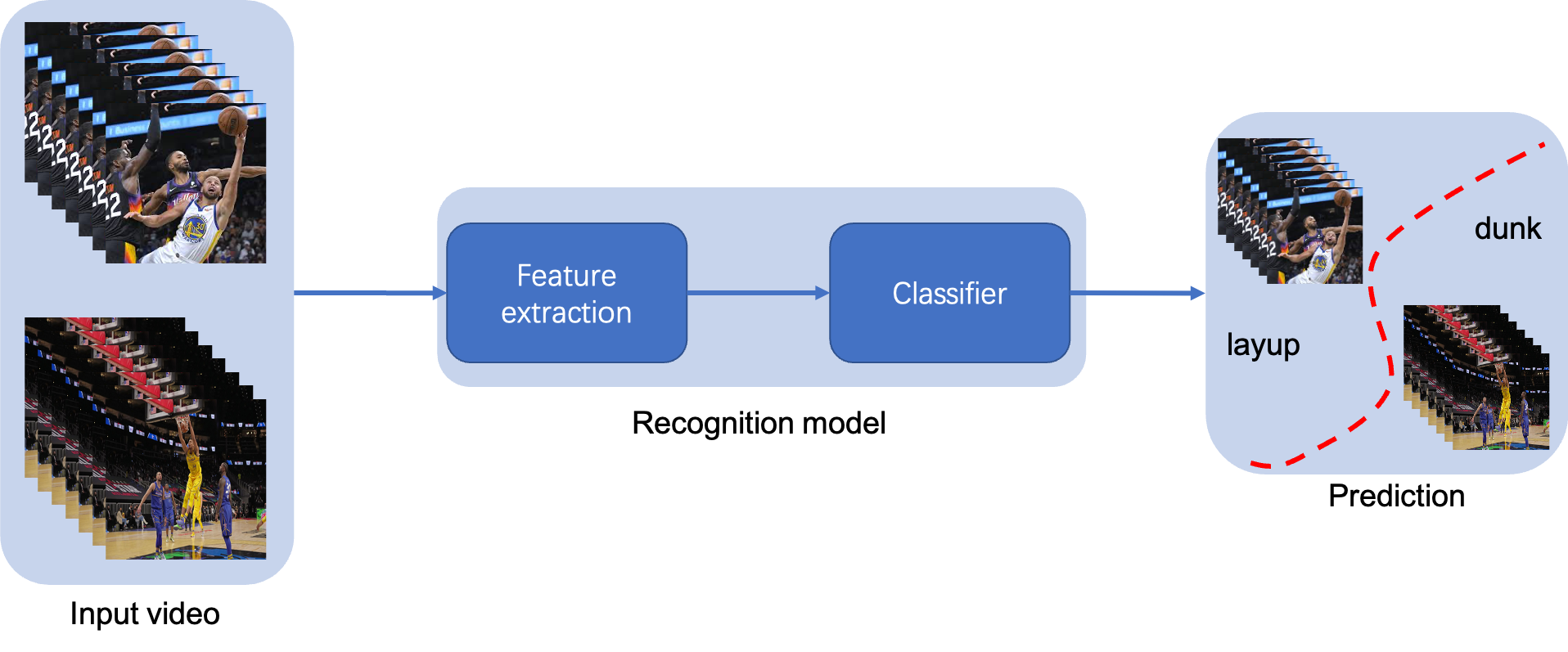}
    \caption{An illustration of action recognition models. Generally, a feature extraction module and a classifier are required for action recognition.}
    \label{fig:action_sys}
\end{figure}

In this section, we dive in to the review of individual action recognition, \ie, each action involves only one person. 

\subsection{Traditional Models}
Generally, an action recognition model consists of at least two modules: (1) video feature extraction and (2) classifier, which is shown in Fig. \ref{fig:action_sys}. Hand-crafted features dominates traditional models. One simple approach is extracting low/middle-level features of each frame using GIST \cite{oliva2001modeling} or \emph{Histogram of Oriented Gradients} (HOGs) \cite{dalal2005histograms} and then averaging the frame features over time for classification \cite{kuehne2011hmdb}. H. Kuehne \etal \cite{kuehne2011hmdb} evaluate multiple feature extraction approaches on various datasets, such as \textbf{UCF Sports} \cite{rodriguez2008action}, showing that using GIST features achieves better performance (60.0\%) than using HOGs (58.6\%) on \textbf{UCF Sports} since the features are biased to the background, for example, the sports of ball normally occur on grass field.

Instead of using 2D HOGs, E. Ijjina \cite{ijjina2020action} applies HOG3D \cite{klaser2008spatio} to extract video features and a \emph{multi-layer perceptron} (MLP) as classifier. In contrast, T. Campos \etal \cite{de2011evaluation} employ HOG3D features $+$ \emph{kernelized Fisher discriminant analysis} (KFDA) for tennis action recognition, achieving AUC of 84.5\% on ACASVA \cite{de2011evaluation}.

Action bank is proposed by S. Sandanand and J. Corso \cite{sadanand2012action}, which is a high-level representation for action recognition, Action bank employs a template-based action detector, which is invariant to appearance changes. The detector is also applied to multi-scale and multi-view videos to be more robust to scales and viewpoints. After that, template actions are selected. Generally, action bank with $N$ action detectors and $M$ samples yields a $N \times M\times 73$-D feature space. Using action bank for feature extraction achieves the accuracy of 95\% on \textbf{UCF sports}.

It is believed that motion plays an important role in action recognition, and various approaches are proposed to use motion information for action recognition, such as \emph{Motion Boundary Histogram} (MBH) \cite{dalal2006human}, \emph{Histograms of Optical Flow} (HOF) \cite{pervs2010histograms} and dense trajectories \cite{wang2013dense}, all of which are based on optical flow. MBH is more robust to camera motion, achieving better performance. H. Wang \etal \cite{wang2013action} propose improved trajectories for action recognition, where camera motion is taken into account, and the model is able to concentrate on the moving objects, achieving much better performance, for example, using the original trajectories achieves the accuracy of 62.4\% on \textbf{Olympic} dataset and MBH achieves 82.4\%, whereas using the improved trajectories finally obtains 91.1\% on \textbf{Olympic} \cite{niebles2010modeling}.

In addition to HOG, \emph{Scale-Invariant Feature Transform} (SIFT) \cite{lowe2004distinctive} is also widely applied to action recognition. M. Chan \etal \cite{chen2009mosift} propose motion SIFT (MoSIFT) to extract video features, where both spatial and temporal are considered, \ie, first, MoSIFT employs histogram of gradients to extract spatial appearance and then employs histogram of optical flow to extract motion features. MoSIFT achieves 89.5\% accuracy on \textbf{Hockey Fight} \cite{bermejo2011violence}, outperforming \emph{Space-Time Interest Points} (STIP) \cite{laptev2005space} (59.0\%).

Though spatial-temporal features extracted using HOG, HOF and SIFT can achieve relatively good performance on sports action recognition datasets like \textbf{UCF Sports} and \textbf{Olympic} (see Table \ref{tab:traditional}), it is normally time-consuming to calculate hand-crafted spatial-temporal features. Moreover, traditional models cannot be trained in a end-to-end manner, \ie, feature extraction module and classifier are learned separately. Recently, researchers pay more attention to deep learning models, proposing many approaches to sports video action recognition and boosting the accuracy of recognition to a higher level.

\begin{table}[t]
\caption{Traditional models for Action Recognition.}\label{tab:traditional}
    \centering
    \scalebox{1}{
    \begin{tabular}{|c|c|c|c|}
        \hline
        Method & Venue & UCF Sports & Olympic \\
        \hline\hline
        Kovashka \emph{et al.}~\cite{kovashka2010learning}&CVPR-2010& 87.27 &-\\
        Wang \emph{et al.}~\cite{wang2013action}&CVPR-2011& 88.20&-\\
        Klaser \emph{et al.}~\cite{klaser2010will}&THESIS-2010& 86.70&-\\
        Wu \emph{et al.}~\cite{wu2011action}&CVPR-2011& 91.30&-\\
        Sadanand \emph{et al.}~\cite{sadanand2012action}&CVPR-2012& 88.20&-\\
        Wang \emph{et al.}~\cite{wang2009evaluation}&BMVC-2009&-&92.10\\
        Laptev \emph{et al.}~\cite{laptev2008learning}&CVPR-2008&-&91.80\\
        Wong \emph{et al.}~\cite{wong2007learning}&CVPR-2007&-&86.70\\
        Schuldt \emph{et al.}~\cite{schuldt2004recognizing}&ICPR-2004&-&71.50\\
        Kim \emph{et al.}~\cite{kim2007tensor}&CVPR-2008&-&95.00\\
        Niebles \emph{et al.}~\cite{niebles2010modeling}&ECCV-2010&-&72.10\\
        \hline
    \end{tabular}
    }
\end{table}

\renewcommand{\thempfootnote}{\fnsymbol{mpfootnote}}
\begin{sidewaystable*}[!htp]
\caption{Deep learning models for Individual Action Recognition.}\label{tab:individual}
    \centering
    \scalebox{0.7}{
    \begin{tabular}{|c|c|c|c|c|c|c|c|c|c|c|c|c|}
        \hline
        \multirow{2}*{Type} & \multirow{2}*{Method\footnote[1]{All the reported methods have been evaluated on at least one sports video dataset or related.}} & \multirow{2}*{Venue} & \multirow{2}*{Pre-train} & \multirow{2}*{Backbone}  & \multicolumn{3}{c|}{Generic} & \multicolumn{5}{c|}{Sports}\\
        \cline{6-13}
        & & & & & Kinetics400 & UCF101 & HMDB51 & Sports1M & FineGym & FSD-10 & P$^2$A & Diving48\\
         \hline\hline
        \multirow{13}*{2D}
        & Slow fusion~\cite{karpathy2014large} & CVPR-2014 &- &- & - &- &- &60.9 &-&-&-&- \\
        &CNN-LSTM \cite{yue2015beyond} & CVPR-2015 & ImageNet & GoogLeNet &- &88.6 &- &73.1 &- &- &- &- \\
        & LRCN \cite{donahue2015long} &CVPR-2015 & ImageNet & AlexNet &- &82.7 &-&-&-&-&-&-\\ 
        & Composite LSTM \cite{srivastava2015unsupervised} &ICML-2015  & ImageNet, Sports1M & VGG-16 &- & 75.8 &44.0 &-&-&-&-&- \\
        & LENN \cite{gan2016you} & CVPR-2016 & - & VGG-16 &- &76.3 &-&-&-&-&-&-\\ 
        & TSN~\cite{wang2018temporal} & TPAMI & ImageNet & ResNet50, BN-Inception &-& 87.3 &-&-&61.4&59.3& 72.1&- \\
        & Attention-LSTM~\cite{long2018attention} & CVPR-2018& ImageNet &  Inception-ResNet-v2, ResNet152 & 79.4 & 94.6 & 69.2 &-&-&-&73.3&-\\
        & TSM~\cite{lin2019tsm} & ICCV-2019& ImageNet & ResNet50 & 74.1 & 95.9 & 73.5 &-&70.6&-& 76.8&-\\
        &KTSN~\cite{liu2020fsd} &arxiv &-&- &-&-&- &-&- &63.3 &-&-\\
        & TimeSformer~\cite{bertasius2021space} & ICML-2021 & ImageNet & ViT-base & 78.0 &-&-&-&-&-& 77.4 & 81.0\\
        & VTN \cite{neimark2021video} & ICCVW-2021 &ImageNet &ViT-base & 79.8 &-&- &-&-&-&-&-\\ 
        \hline\hline
        \multirow{19}*{3D}
        & C3D~\cite{tran2015learning} & ICCV-2015 & Sports1M & VGG16 & 59.5 & 82.3 & 56.8 & 61.1 &-&-&-&-\\
        & I3D~\cite{carreira2017quo} & CVPR-2017 & ImageNet, Kinetics & BN-Inception & 71.1 & 95.6 & 74.8 &-& 63.2 &-&-&-\\
        & P3D~\cite{qiu2017learning} & ICCV-2017 & Sports-1M & ResNet50 & 71.6 & 88.6 & - &-&-&-&-&-\\
        & R(2+1)D-RGB~\cite{tran2018closer} & CVPR-2018 & Sports1M, Kinetics & R3D-34 & 72.0 & 96.8 & 74.5 & 73.0 &-&-&-&-\\
        &S3D \cite{xie2018rethinking} &ECCV-2018 &ImageNet, Kinetics & BN-Inception &74.7 & 96.8 &- &-&-&-&-&-\\ 
        &CSN~\cite{tran2019video} & ICCV-2019 & IG-65M~\cite{ghadiyaram2019large} & R3D-152& 82.6 &-&-& 75.5 &-&-&-&-\\ 
        & SlowFast~\cite{feichtenhofer2019slowfast} &ICCV-2019 &-& ResNet101 & 79.8 &-&-&-&-&-&77.4&77.6\\
        &STM \cite{jiang2019stm} &ICCV-2019 &ImageNet, Kinetics &ResNet50 &73.7 &96.2 &72.2 &-&-&-&-&-\\
        & X3D~\cite{feichtenhofer2020x3d} &CVPR-2020 &-&- &79.1 &-&- &-&-&-&-&-\\ 
        & TPN~\cite{yang2020temporal} &CVPR-2020 &- &ResNet101 &79.8 &-&- &-&-&-&-&-\\ 
        & ViViT~\cite{arnab2021vivit} &CVPR-2021 &- &ViViT-large &81.3 &-&- &-&-&-&-&-\\ 
        &MViT \cite{fan2021multiscale} & CVPR-2021 &- &MViT-base &81.2 &- &- &-&-&-&-&-\\ 
        & MoViNet~\cite{kondratyuk2021movinets} & CVPR-2021&-& MoViNet-v6 & 81.5 &-&-&-&-&-& 74.1 &-\\
        & ViSwin~\cite{liu2021video} & arXiv & ImageNet & ViSwin-large & 84.9 &-  &-&-&-&-&81.0&-\\
        & ORViT TimeSformer~\cite{herzig2021object} & arXiv & ImageNet & ORViT&-&-&-&-&-&-&-&88\\
        & BEVT~\cite{wang2021bevt} & arXiv & ImageNet, Kinetics & ViSwin-base &80.6&-&-&-&-&-&-&86.7\\
        & MaskFeat \cite{wei2021masked} & arxiv & Kinetics & MViT-large &87.0 &-&- &-&-&-&-&-\\ 
        & VIMPAC~\cite{tan2021vimpac} & arXiv & HowTo100M~\cite{miech2019howto100m} & BERT-L~\cite{devlin2018bert} &77.4&92.7&65.9&-&-&-&-&85.5\\
        & TFCNet~\cite{zhang2022tfcnet} & arXiv & ImageNet & R3D-50 &-&-&-&-&-&-&-&88.3\\
        \hline\hline
        \multirow{8}*{Two-stream} 
        & Two-Stream ConvNet~\cite{simonyan2014two} &NIPS-2014 &-&- &- &88.0 &59.4 &-&-&-&-&-\\
        &Two-Stream Fusion~\cite{feichtenhofer2016convolutional} &CVPR-2016 &-&VGG-16 &- &92.5 &65.4 &-&-&-&-&-\\
        & TSN-Two-Stream~\cite{wang2018temporal} & ECCV-2016 & ImageNet & ResNet50, BN-Inception & 73.9 & 94.0 & 68.5 &-&76.4&76.0& 72.1&-\\
        &R(2+1)D-Two-Stream~\cite{tran2018closer} & CVPR-2018 & Sports1M, Kinetics & R3D-34 & 75.4 & 97.3 & 78.7 & 73.3 &-&-&-&-\\
        & TRN-Two-Stream~\cite{zhou2018temporal} & ECCV-2018& ImageNet & BN-Inception & 63.3 & 83.8 &-&-&79.8&-&-&-\\
        & TSM-Two-Stream~\cite{lin2019tsm} & ICCV-2019& ImageNet & ResNet50 &-&-&-&-&81.2&-&-&-\\
        & KTSN-Two-Stream~\cite{liu2020fsd} & arXiv & ImageNet & ResNet50, BN-Inception &-& 94.9 & 82.1 &-&-&82.6&-&-\\
        & G-Blend~\cite{wang2020makes} & CVPR-2020& IG-65M & R3D-50 & 83.3 &-&-&62.8 &-&-&-&-\\
        \hline\hline
        \multirow{5}*{Skeleton} &
        ST-GCN~\cite{yan2018spatial} & AAAI-2018 &-&GCN& 30.7\footnote[2]{The performances of all the skeleton-based algorithms are conducted on the Kinetics-Skeleton-400 dataset.} &-&-&-&25.2&60.5&-&-\\
        & AGCN~\cite{shi2020skeleton} & TIP &-&GCN&36.1$^\dagger$ &-&-&-&-&65.9&-&-\\
        & EfficientGCN~\cite{song2020stronger} &MM-2020&-&GCN&-&-&-&-&-&65.5&-&-\\
        & CTR-GCN~\cite{chen2021channel} &ICCV-2021&-&GCN&-&-&-&-&-&66.2&-&-\\
        & PoseC3D~\cite{duan2021revisiting} &CVPR-2022&-&C3D& 47.7$^\dagger$ &-&-&-&94.3&68.8&-&-\\
        \hline
    \end{tabular}
    }
\end{sidewaystable*}

\subsection{Deep Models}\label{sec:deep-model}

\begin{figure}[t]
    \centering
    \includegraphics[width=0.8\linewidth]{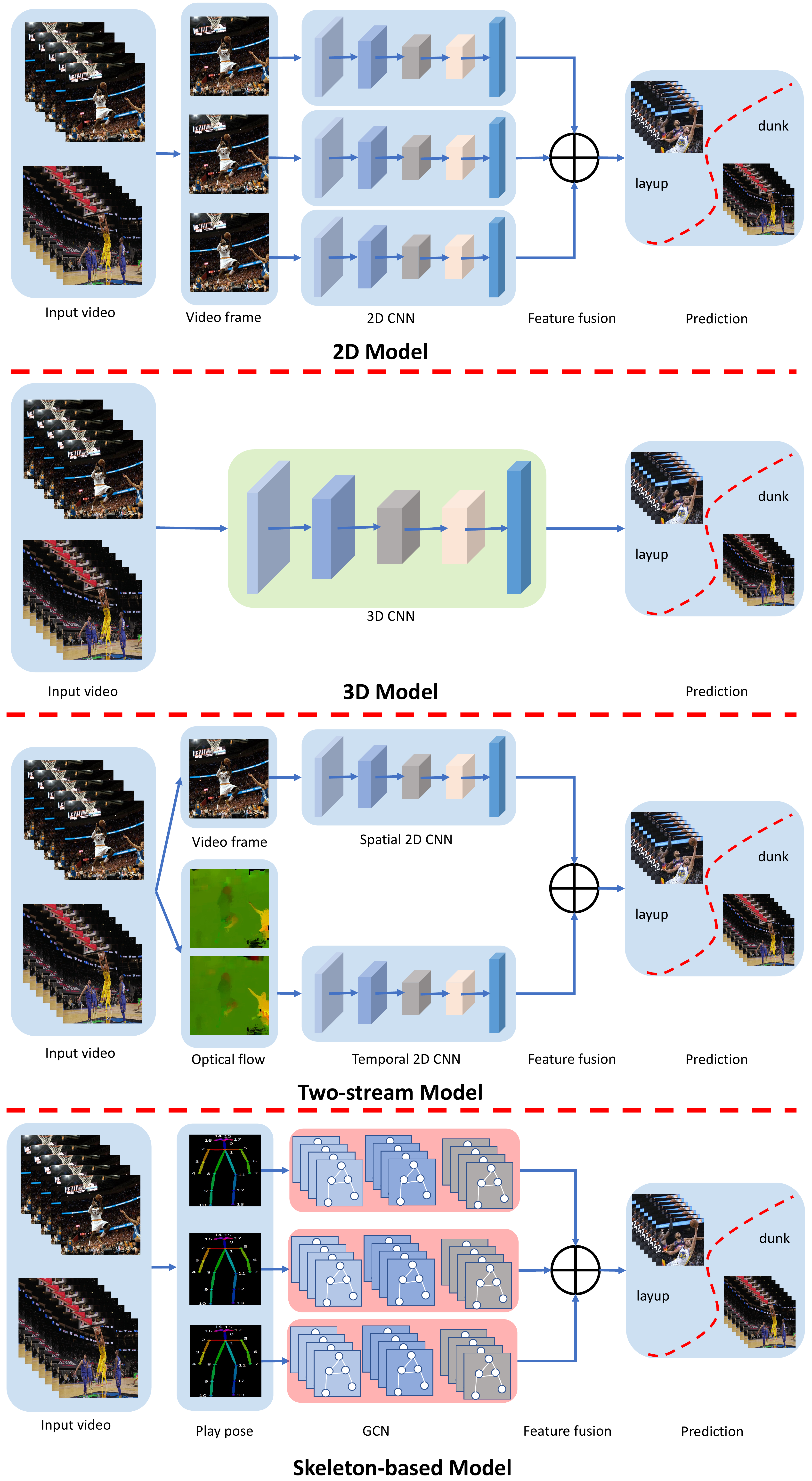}
    \caption{An illustration of deep models for action recognition. We present 4 types of deep models: \textbf{2D model}, \textbf{3D model}, \textbf{two-stream model} and \textbf{skeleton-based model}. Note that we only present the basic frameworks of these models and there could be some other variants (more details can be found in section \ref{sec:deep-model}).}
    \label{fig:deep-models}
\end{figure}

Currently, deep models dominate video action recognition. Compared with traditional models, deep models are more feasible and can be trained in a end-to-end manner. Typically, there are four types of deep models: 2D model, 3D model, Two/multi-stream model and skeleton-based model. We show the basic architectures of four typical models in Fig. \ref{fig:deep-models} and more details can be found in the following subsections.

\subsubsection{\textbf{2D Models}}
2D models employ 2D convolutional neural networks (CNN) or transformers \cite{dosovitskiy2020image} to process each video frame separately and then fuse the extracted features for prediction.

A. Karpathy \etal \cite{karpathy2014large} introduce CNNs into video action recognition, proposing four time information fusion approaches: (1) single-frame fusion -- using a shared CNN to extract features of each single frame and then concatenate the final representations for classification, (2) early fusion -- using a 3D kernel with the size of $11 \times 11 \times 3 \times T$ to combine information of frames across a time window, (3) late fusion -- using a shared CNN to compute the representations of two separate frames with the distance of 15 frames and a fully connected layer to fuse the single-frame representations (4) slow fusion -- implementing a 3D kernel in the first layer and then slowly fusing the information of frames in higher layers of the network. The experiments show that slow fusion is superior to other fusion approaches, for example, slow fusion obtains 60.9\% accuracy on \textbf{Sports 1M} \cite{karpathy2014large}, while single-frame fusion, early fusion and late fusion achieve 59.3\%, 57.7\% and 59.3\%, respectively. Interestingly, using hand-crafted features like HOG only achieves 55.3\% accuracy, which is considerably lower than using CNNs, indicating that deep models are promising for sports video action recognition and inspiring researchers to develop more deep models.

Another family of 2D deep models is directly using \emph{Long-short Term Memory} (LSTM) \cite{hochreiter1997long} networks to capture temporal information, which is relatively popular in early deep models. In 2015, Y. Ng \etal \cite{yue2015beyond} propose an approach that combines 2D CNNs and LSTMs, \ie, first, using a shared 2D CNN to obtain spatial representations of frames and then then applying a multi-layer LSTM to fuse the spatial representations. Also, J. Donahue \etal \cite{donahue2015long} propose a similar model which uses a two-layer LSTM, termed \emph{Long-term Recurrent Convolutional Networks} (LRCN). While N. Srivastava \etal \cite{srivastava2015unsupervised} employ LSTM-based auto-encoder to learn better video representations trained in an unsupervised manner. Latter, C. Gan \etal \cite{gan2016you} propose a \emph{Lead-exceed Neural Network} (LENN) which is similar to the model in \cite{yue2015beyond}, but LENN uses web images to fine-tune the lead network to filter out irrelevant video frames. 

As mentioned above, temporal information fusion is crucial in 2D models. Alternatively, L. Wang \etal \cite{wang2018temporal} propose a \emph{Temporal Segment Network} (TSN) for video action recognition, which is composed of a spatial CNN and a temporal CNN. First, an input video is divided into some segments and the short snippets composed of RGB frames, optical flow and RGB differences are randomly sampled from segments. After that, the snippets are fed into spatial and temporal networks to make predictions. Finally, we can obtain a prediction via aggregating the snippet prediction scores. TSN uses temporal information in two ways: (1) it directly introduces optical flow into the framework, (2) similar to late fusion in \cite{karpathy2014large}, TSN aggregates the snippet predictions. Finally, the 2D TSN that only using RGB frames obtains impressive performance, for example, 61.4\% accuracy on \textbf{FineGym} \cite{shao2020finegym} and 87.3\% on the generic action recognition dataset -- \textbf{UCF101} \cite{soomro2012ucf101}. Another variant of TSN is using key video frames instead of random sampling, namely KTSN \cite{liu2020fsd}. Applying key video frames achieves better performance on \textbf{FSD-10}, \ie, 63.3\% vs. 59.3\%.

Instead of using simple aggregation approaches, such as concatenation and linear combination, B. Zhou \etal \cite{zhou2018temporal} propose a \emph{Temporal Relational Network} (TRN) to capture the temporal relations among frames, where the relations are computed using a MLP and can be plugged into any existing frameworks. TRN remarkably improves the performance on \textbf{FineGym} \cite{shao2020finegym}, obtaining 68.7\% accuracy.

However, using MLPs in TRN is time-consuming when considering many frames and cannot well capture useful low-level features. To address this issue, J. Lin \etal \cite{lin2019tsm} propose a simple yet efficient module, namely \emph{Temporal Shift Module} (TSM) to capture temporal information for action recognition, where spatial features are extracted using 2D CNNs on video frames and then inserting TSM into 2D convolutional blocks. TSM achieves 70.6\% accuracy on \textbf{FineGym} \cite{shao2020finegym}, outperforming 2D TSN, 2D TRN and some 3D approaches like I3D \cite{carreira2017quo} but having lower computational complexity.

In recent 2 years, vision transformers (ViT) \cite{dosovitskiy2020image} become increasingly popular for computer vision tasks, where multi-head self-attention \cite{vaswani2017attention} is employed to replace convolutional kernels. G. Bertasius \etal \cite{bertasius2021space} investigate different combinations of spatial self-attention and temporal self-attention (space-only, joint space-time, divided space-time, sparse local-global and axial attention), where spatial attention is performed over patches belong to the same video frame and temporal attention is applied to patches across frames, yielding a model termed TimeSformer. Experiments show that using divided space-time attention outperforms other architectures, achieving 81.0\% accuracy on \textbf{Diving48} \cite{li2018resound}. While \emph{Vision Transformer Network} (VTN) \cite{neimark2021video} employs a temporal transformer to fuse frame representations, obtaining 79.8\% accuracy on \textbf{Kinetics-400} \cite{kay2017kinetics}.

In summary, for 2D deep models, we can find that both spatial and temporal modules are shifting to transformers since transformers are much more powerful to model sequences and extract frame features, however, transformers have more learnable parameters, requiring more computational resources. In addition, training a large model is non-trivial due to the difficulty of convergence. Another trend is adopting pre-training, \ie, using large-scale image dataset like ImageNet \cite{deng2009imagenet} to pre-train the spatial networks.

\subsubsection{\textbf{3D Models}}
Compared with 2D models, 3D models normally treat a sequence of frames as a whole and apply 3D convolutional neural networks or cube-based transformers to simultaneously capture spatial and temporal information.

3D CNN for action recognition \cite{ji20123d} is a pioneer work proposed by S. Ji \etal, which is composed of a hardwired layer, two 3D convolutional layers, two subsampling layers, one 2D convolutioinal layer and a fully-connected layer. Though the proposed network is relatively small and only evaluated on small datasets, this work presents a prototype of 3D CNNs for action recognition and achieves better performance than using 2D CNNs.

Later, in 2015, D. Tran \etal \cite{tran2015learning} design a modern and deep 3D architecture -- C3D for large-scale action recognition, where eight 3D convolutional layers with $3\times 3\times 3$ kernel size are adopted. C3D obtains 61.1\% accuracy on \textbf{Sports 1M} \cite{karpathy2014large}, which is relatively competitive. Likewise, J. Carreira and A. Zisserman \cite{carreira2017quo} propose a \emph{Inflated 3D CNN} (I3D), where a 2D kernel with $N\times N$ size is expanded into a $N\times N\times N$ 3D kernel and the parameters of 3D kernels are also from pre-trained 2D kernels via bootstrapping. Compared with C3D, I3D is much deeper, stacking 9 3D inception modules \cite{pouyanfar2017efficient} and 4 individual 3D comvolutional layers. With these modern designs, I3D obtains much better performances on multiple datasets, for example, 95.6\% vs. 82.3\% on \textbf{UCF101} \cite{soomro2012ucf101}.

Directly expanding $N\times N$ 2D convolution into $N\times N \times N$ 3D convolution can significantly increase the number of parameters, improving the capacity of deep models but also raising computational complexity and the risk of overfitting. To mitigate address the problem, Z. Qiu \etal \cite{qiu2017learning} propose a \emph{Pseudo 3D} (P3D) network, where 3D convolution is substituted by stacking a 2D convolution and an 1D convolution. Similarly, D. Tran \etal \cite{tran2018closer} explores different architectures (2D, 3D and (2+1)D), finding that stacking a 2D convolution with $1\times N \times N$ kernel size and a $t\times 1\times 1$ 1D convolution is superior to other architectures. While S3D \cite{xie2018rethinking} replaces part of 3D inception modules in I3D \cite{carreira2017quo} with 2D inception modules to balance the performance and computational complexity. Later, D. Tran \etal \cite{tran2019video} propose set of architectures, termed \emph{– 3D Channel-Separated Networks} (CSN), to further reduce FLOPs, where group convolution, depth convolution and different combinations of then are explored. CSN achieves much better performance than 3D CNNs with only one third FLOPs of 3D CNNs. 

SlowFast \cite{feichtenhofer2019slowfast} is composed of two branches -- one is the slow branch with low frame rate and another is the fast branch with high frame rate. The slow branch with low frame rate can pay more attention to spatial semantics, while the fast branch pays more attention to object motion. To achieve this, the network of slow branch is designed only using 2D convolution in the bottom layers and using (1+2)D convolution in the top layers, whereas the fast branch uses (1+2)D convolution in each layer. Note that the fast branch is designed to capture object motion instead of high-level semantics, thus it can be a lightweight neural network. In addition, SlowFast adopts lateral connections to fuse slow and fast features. With elaborate designs of slow branch, fast branch and lateral connections, SlowFast achieves state-of-the-art performance on several popular action recognition datasets.

To model long video sequences, S. Zhang \cite{zhang2022tfcnet} introduces \emph{Temporal Fully Connected Operation} into SlowFast, proposing TFCNet, where the features of all frames are combined by a FC layer. Whith a simple operation, TFCNet boosts the performance on \textbf{Diving48} to 88.3\%, nearly 11\% higher than that achieved by SlowFast.

STM \cite{jiang2019stm} adopts two modules -- \emph{Channel-wise Spatial-Temporal Module} (CSTM) and \emph{Channel-wise Motion Module} (CMM), where CSTM employs (2+1)D convolution to fuse spatial and temporal features, while CMM only uses 2D convolution but concatenates the features of three successive frames. Compared with P3D \cite{qiu2017learning} and R3D \cite{tran2018closer}, STM performs better.

X3D \cite{feichtenhofer2020x3d} expand 2D CNNs in four manners -- space, time, depth and width, which explores a number of architectures, finding that high spatial-temporal networks is superior to other models. X3D is inferior to SlowFast on \textbf{Kinetics-400} (79.1\% vs. 79.8\%), but X3D has fewer parameters and takes less time during training and inference. To further reduce the number of parameters and FLOPs, D. Kondratyuk \etal \cite{kondratyuk2021movinets} propose \emph{Mobile Video Networks} (MoViNets) that are able to process streaming videos. Tow core techniques are applied in MoViNets -- the first one is \emph{Neural Architecture Search} (NAS) \cite{bender2020can} for  efficient architectures generation and the second one is stream buffer technique that equips 3D CNNs to tackle streaming videos with arbitrary length. With these two techniques, MoViNets only requires 20\% FLOPs of X3D, but achieves better performance. 

SlowFast \cite{feichtenhofer2019slowfast} shows that introducing different temporal resolutions benefits action recognition, however, it applies an individual network to each resolution, which is time-consuming. In contrast, TPN \cite{yang2020temporal} applies one backbone network and uses temporal pyramid to 3D features in different levels, \ie, low frame rate for high-level feature to capture semantics and high frame rate in low-level features to capture motion information. TPN achieves the same performance on \textbf{Kinetics-400} but only adopts one branch.

After 2020, the number of transformers using 3D modules is rising. Compared with 2D transformer-based models like TimeSformer \cite{bertasius2021space} which separately uses spatial and temporal self-attention, 3D transformer-based models execute self-attention over non-overlap cubes, which is more similar to 3D convolution. ViViT \cite{arnab2021vivit} expand ViT into video action recognition via using tubelet embedding. Also, ViViT explores different architectures of transformers -- spatial-temporal transformer, factorised encoder, factorised self-attention and factorised dot-product, finding that spatial-temporal transformer performs the best on large datasets but overfits small datasets and needs much more FLOPs than other architectures since spatial-temporal transformer executes self-attention over all tokens with a computational complexity of $N_t^2$, where $N^2_t$ denotes the number of tokens.

MViT \cite{kondratyuk2021movinets} mimic the multi-scale architectures of CNNs, introducing multi-head pooling attention into ViT \cite{dosovitskiy2020image}, \ie, high resolution for low-level features and low resolution for high-level features. In terms of action recognition, 3D pooling attention is applied. Though MViT executes self-attention over all spatial-temporal tokens, the number of tokens drops when it goes deeper and the dimension of token embedding is low in shallow layers, hence, the FLOPs of MViT is around 1/5 of ViViT FLOPs. Compared with ViViT, MViT with fewer parameters and less computational cost  achieves similar performance on \textbf{Kinetics-400}.

Similar to MViT, \emph{Video Swin Transformer} (ViSwin) \cite{liu2021video} uses different resolutions in different levels, but it only reduces the spatial resolution in each level and keeps the temporal resolution. One important property of ViSwin is using 3D shifted window based self-attention, which reduces the computational complexity and increases the receptive field via stacking multiple layers. Finally, ViSwin-large achieves 84.9\% accuracy on \textbf{Kinetics-400} with ImageNet-21K pre-trained parameters and a high spatial resolution (384$\times 384$).

As we have mentioned above, transformer-based models normally split frames into 2D non-overlap patches or 3D non-overlap cubes, thus, the objects in videos could be divided into different patches or cubes. missing object-centric information. ORViT, short for \emph{Object-Region Vision Transformer} \cite{herzig2021object} introduces object-dynamic module and object-region attention into vision transformers. In object-dynamic module, object bounding box coordinates are encoded using box position encoder, while in object-region attention module, object representations obtained by RoIAlign \cite{he2017mask} are employed to generate key and value vectors. With these two modules, ORViT pays more attention to objects and achieves 88\% accuracy on \textbf{Diving48}, 8\% higher than the baseline. Though introducing object features can benefit the model to capture more semantics, it requires multi-object tracking to obtain the bounding boxes of objects.

Similar to \emph{Masked Language Models} (MLM) \cite{devlin2018bert}, researchers also develop a number of masked video models. BEVT \cite{wang2021bevt} expands BEIT \cite{bao2021beit} to video domain. Briefly, BEVT predicts the representations of masked patches, where the presentations are obtained by VQ-VAE \cite{ramesh2021zero}. Likewise, VIMPAC \cite{tan2021vimpac} predicts patch representations obtained by VQ-VAE, but uses a 24-layer BERT-like backbone instead of ViSwin \cite{liu2021video} and applies contrastive learning during training -- discriminating positive video clip pairs from negative ones. Though VIMPAC employs both patch representation prediction and contrastive learning, it is inferior to BEVT and one possible reason is that ViSwin is more powerful and the parameters of the image Swin are shared with ViSwin, hence, it can well model spatial information. Alternatively, MaskFeat \cite{wei2021masked} employs MViT \cite{kondratyuk2021movinets} as the backbone and explores predicting the features of the masked patches obtained by different approachs, such as HOG, VQ-VAE and DINO \cite{caron2021emerging}, finding that predicting HOG is slightly worse than using DINO but DINO requires a pre-trained model. 

Through the numbers in Table \ref{tab:individual}, we can make the conclusion that 3D models are normally superior to 2D models, but 3D models could be time-consuming and cost more computational resources. Also, we can find that pre-train-fine-tune paradigm is increasingly popular for 3D models, in particular for 3D transformer-based models since it is straightforward to introduce the tricks of MLM into video models.

\subsubsection{\textbf{Two-stream Models}}
Two-steam models normally take RGB frames and optical flow as input and each stream employs a deep neural network (see Fig. \ref{fig:deep-models}). RGB frames provide both spatial and temporal information, while optical flow mainly provides information of motion. Obviously, we can easily expand the above 2D/3D models that only take RGB frames as input into two-stream models, resulting in their two-streams variants, such as TSN-Two-Stream \cite{wang2018temporal}, TSM-Two-Stream \cite{lin2019tsm} and TRN-Two-Stream \cite{zhou2018temporal}. Compared with their one-stream versions that only use video frames, two-stream models achieve better performance but require to calculate optical flow first and an additional neural network to obtain deep representations of motion.

Another problem of two-stream models is that how to combine the representations of frames and optical flow. An early work Two-Stream ConvNet \cite{simonyan2014two} proposed by K. Simonyan \etal directly average the prediction of each stream, while C. Feichtenhofer \etal \cite{feichtenhofer2016convolutional} explores different fusing approaches, including max-pooling, concatenation, bilinear, sum and convolution in different layers of the two stream networks. 

Recently, researchers observe that some advanced one-stream models outperform its two-stream counterparts since tow-stream networks have higher capacity, easily overfitting the dataset. In addition, the generalizabilities  of using video frames and optical flow are different, so training two-stream network with one strategy is sub-optimal. W. Wang \etal \cite{wang2020makes} endeavours to address the issues, proposing \emph{Gradient Blending} (G-Blend) where the weights of different loss functions are estimated during training, hence, it assigns a weight to each stream.

\subsubsection{\textbf{Skeleton-based Models}}

2D, 3D and two-stream deep models take RGB frames as input, while skeleton-based models take players' skeleton graph as input (see Fig. \ref{fig:deep-models}). Normally, \emph{Graph Convolutional Networks} (GCN) \cite{kipf2016semi} are used to model the skeleton graph composed of joints.

S. Yan \etal~\cite{yan2018spatial} propose a \emph{Spatial-Temporal GCN} (ST-GCN) for action recognition, which is similar to 3D convolutional networks but executed on skeleton graph, achieving 30.7\% accuracy on \textbf{Kinetics-400}. Compared with frame based models like 2D and 3D models, the performance of ST-GCN is much worse since it cannot capture the appearance information, however, convolution on graphs is much faster.

AGCN~\cite{shi2020skeleton} introduce attention mechanism into GCN. Three types of attention are employed in AGCN -- spatial attention, temporal attention and channel attention. With these types of attention, AGCN achieves higher accuracy scores. Similarly, C. Si \etal \cite{si2019attention} propose an \emph{Attention Enhanced Graph Convolutional LSTM Network} (AGC-LSTM), where the temporal information is captured using a LSTM and the spatial information is captured using a GCN with attention.

Y. Song \etal improve GCNs with a bag of advanced techniques, such as batch normalization \cite{ioffe2015batch}, yielding an EfficientGCN~\cite{song2020stronger} that achieves competitive performance on \textbf{FSD-10}, but takes less time for training and is more explainable.

The topology of graphs is crucial for action recognition and Y. Chen \etal propose a \emph{Channel-wise Topology Refinement GCN} (CTR-GCN)~\cite{chen2021channel} to effectively model the topology. Specificly, CTR-GCN employs channel-wise topology modeling block to compute the channel-wise correlation and then models the relationship among graph nodes in different channels. Finally, CTR-GCN achieves 66.2\% accuracy on \textbf{FSD-10}, better than ST-GCN and AGCN.

The drawback of using skeleton graphs composed of joints is that we need to detect the joints first and normally the predicted graphs are noisy, leading to worse performance on existing datastes. Alternatively, PoseC3D \cite{duan2021revisiting} applies the heatmaps of joints and limbs instead of graphs, which are more robust than directly using skeleton graphs. Pose3D treats the heatmaps as frames, hence, traditional 3D convolutional networks can be adopted. Through Table \ref{tab:individual}, we can find that Pose3D is superior to other skeleton-based models, but still inferior to two-stream models.

As we have mentions above, skeleton-based models require to detect the joints first, resulting in extra computation cost and prediction noise. Though using heatmaps can mitigate the problem of noise, the performance is still worse than other types of models. 

\begin{table}[t]
\centering
\caption{Current state of individual sports video action recognition. Here we only list the performance on the sports-related datasets not in Table \ref{tab:individual}.}\label{tab:sport_state}
\scalebox{0.75}{
\begin{tabular}{|c|c|c|c|c|}
\hline
     Sports & Dataset & Model &Year & Performance  \\ \hline\hline
     \multirow{3}*{Tennis} 
     &ACASVA~\cite{de2011evaluation} &HOG3D+CNN~\cite{ijjina2020action} &2020 & 93.78 \\ \cline{2-5}
     &THETIS~\cite{gourgari2013thetis} &Lightweight 3D~\cite{rasmussen2022compressing} &2022 &90.9 \\ \cline{2-5}
     & TenniSet \cite{faulkner2017tenniset} &Two-stream \cite{faulkner2017tenniset} &2017 &81.0 \\ 
     \hline\hline
     \multirow{3}*{Table tennis} &TTStroke-21~\cite{martin2018sport} & Two-stream \cite{martin2020fine} & 2020 & 91.4 \\ \cline{2-5}
     & SPIN \cite{schwarcz2019spin} & Multi-stream \cite{schwarcz2019spin} &2019 &72.8 \\ \cline{2-5}
     &Stroke Recognition \cite{kulkarni2021table} &TCN \cite{lea2017temporal} & 2021 & 99.37 \\
     \hline \hline
     Badminton 
     & Badminton Olympic \cite{ghosh2018towards} &TCN \cite{lea2017temporal} & 2018 & 71.49 \\ 
     \hline\hline
     \multirow{3}*{Basketball}
     & NCAA \cite{ramanathan2016detecting} & CNN+LSTM \cite{ramanathan2016detecting} &2016 &51.6 \\ \cline{2-5}
     & FineBasketball \cite{gu2020fine}  & TSN-Two-Stream \cite{wang2018temporal} & 2020  & 29.78 \\ \cline{2-5}
     & NPUBasketball \cite{ma2021npu} & Skeleton-based \cite{ma2021npu} & 2020  & 80.9 \\
     \hline\hline
     Football 
     &SoccerNet \cite{giancola2018soccernet} & 3D \cite{giancola2018soccernet} & 2018 & 65.2 \\
     \hline\hline
     \multirow{3}*{Others}
     & Hockey Fight \cite{bermejo2011violence} & Two-stream \cite{zhou2017violent} & 2017 & 97.0 \\ \cline{2-5}
     & GolfDB \cite{mcnally2019golfdb} & CNN+LSTM \cite{mcnally2019golfdb} & 2019 & 79.2 \\ \cline{2-5}
     & FenceNet \cite{zhu2022fencenet} & TCN \cite{lea2017temporal} & 2022 & 87.6 \\
\hline
\end{tabular}
}
\end{table}

\subsubsection{\textbf{Others}} In addition to 2D, 3D, two-stream and skeleton-based models, hybrid models that composed of multiple model types are also applied for video action recognition. One recent work -- \emph{Temporal Query Networks} (TQN) \cite{zhang2021temporal} combines 3D CNNs and transformers. Specifically, 3D CNNs are used as the backbone to extract video features and transformers are adopted as decoders, \ie, given a query, the transformers output a response, where the queries are texts like \texttt{the number of flips} for diving and the responses are the corresponding attributes, such as a number or a label. The transformer-based decoder models the relevance among visual features, queries and responses. In terms of fine-grained action recognition, TQN requires to pre-defined action labels and each label has a set of attributes for classification, hence, we can classify the actions based on the responses. Compared with its 3D counterparts, TNQ shows its superiority, achieving 89.6\% on \textbf{FineGym} and 81.8\% on \textbf{Diving48}.

Note that videos are composed of not only frames but also audios, and there a family of models that adopt multiple modalities. Similar to two-stream models, multimodal models consists of several branches. One recent work is AudioSlowFast \cite{xiao2020audiovisual} proposed by F. Xiao \etal, where acoustic information is introduced into the original SlowFast \cite{feichtenhofer2019slowfast} model using an audio branch, hence, AudioSlwoFast has 3 branches -- slow, fast and audio. While Y. Bian \etal \cite{bian2017revisiting} propose an ensemble model that adopts video frames, optical flow and audio. In our developed toolbox \footnote{https://github.com/PaddlePaddle/PaddleVideo}, we also adopt acoustic information to classify football actions, where there are 8 categories, such as red card, corner and free kick. Using multiple modalities is able to improve the capacity of deep models and the redundant information could make the model more robust, however, it is difficult to combine different modalities and training multimodal models is non-trivial \cite{wang2020makes}. In addition, using more branches leads to a large models, so overfitting can easily occur.

In Table \ref{tab:sport_state}, we present current state of action recognition in different types of sports. We can see that 3D and two-stream models are relatively popular and the recent advanced models like MoViNet~\cite{kondratyuk2021movinets} are rarely used in sports. One possible reason is that some sports-related datasets lack challenges and two-stream models can achieve high accuracy, for example, 91.4\% on \textbf{TTStroke-21} \cite{martin2018sport}. While some other datasets like \textbf{NCAA} \cite{ramanathan2016detecting} and \textbf{FineBasketball} \cite{gu2020fine} are still challenging, requiring more advanced models.

\section{Group/Team Activity Recognition}\label{sec:team}

\begin{figure}[t]
    \centering
    \includegraphics[width=0.8\linewidth]{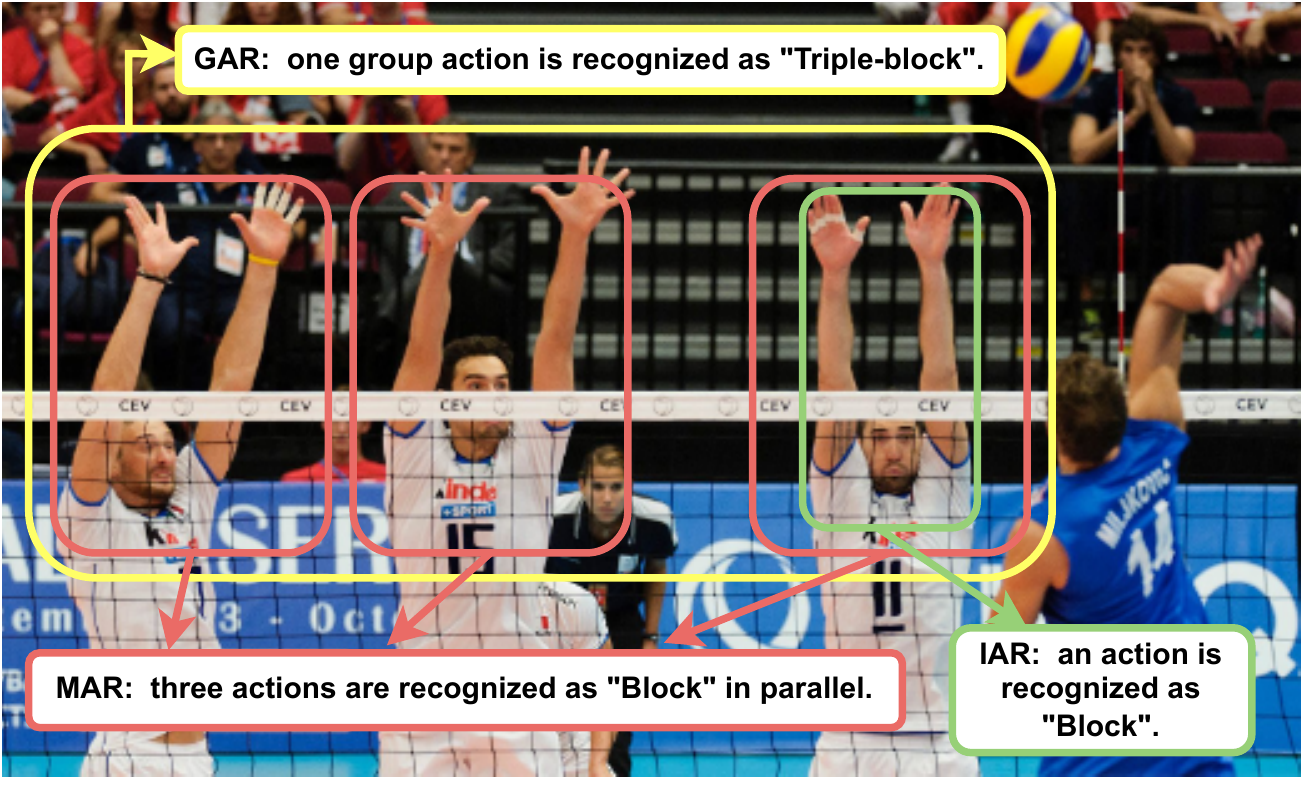}
    \caption{An example of individual, group, and multi-player activity recognition in a frame of volleyball competition video.}
    \label{fig:def_GAR}
\end{figure}

\begin{figure}[t]
    \centering
    \includegraphics[width=0.8\linewidth]{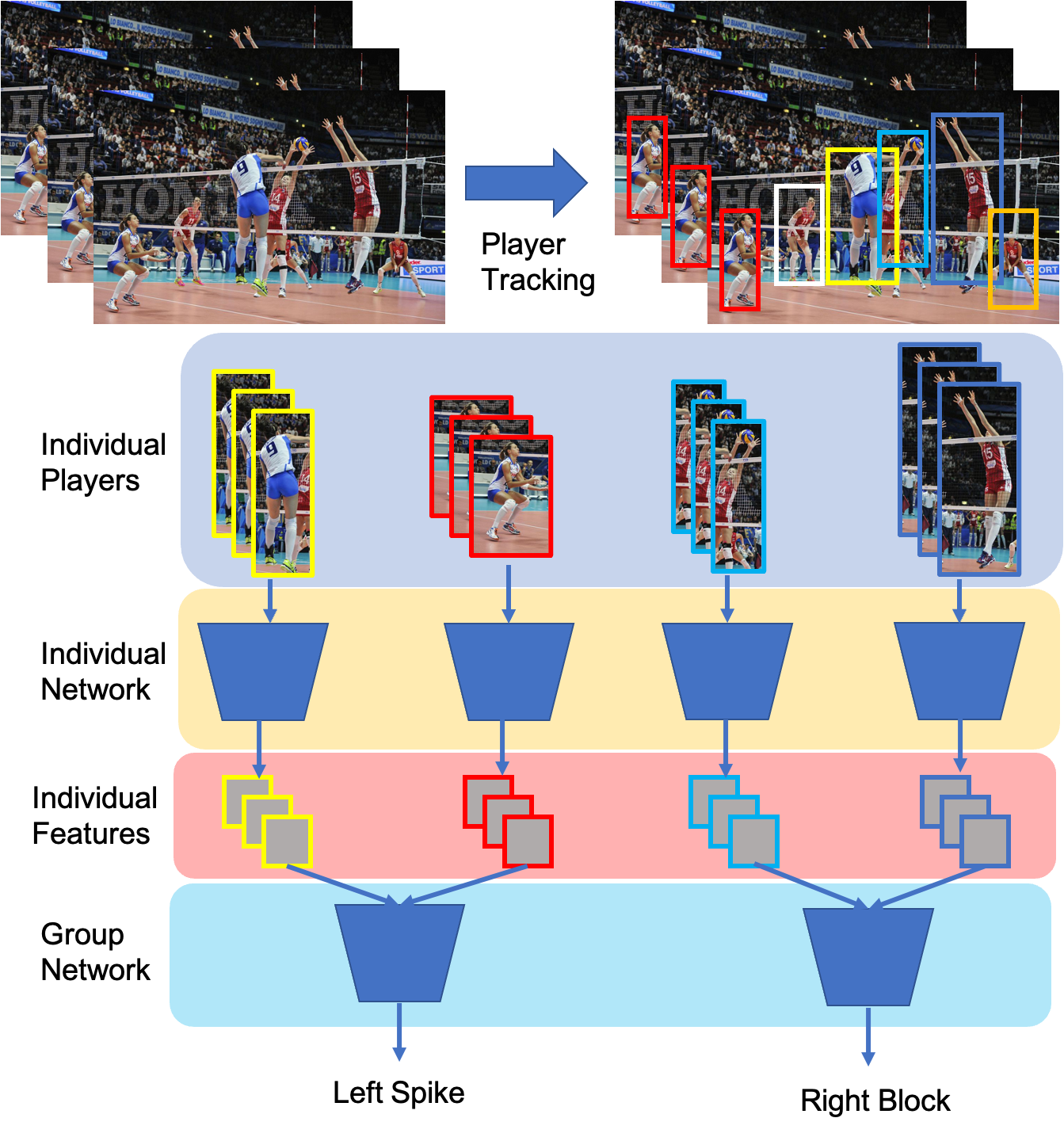}
    \caption{A typical framework for group activity recognition (GAR). Compared with models for individual action recognition shown in Fig. \ref{fig:deep-models}, GAR models normally require player tracking, individual player feature extraction and group feature combination, which is more complicated.}
    \label{fig:gar_framework}
\end{figure}

Group/team activity recognition is one branch of human activity recognition problem which targets the collective behavior of a group of people, resulted from the individual actions of the persons and their interactions. It is a basic task for automatic human behavior analysis in many areas, such as \textbf{sports}, health care and surveillance. Note that, although group/team activity is conceptually an activity performed by mutiple people or objects, the group/team activity recognition (GAR) is quite different from another common task -- the multi-player activity recognition (MAR)~\cite{gordon2014group}. The former is the process of recognizing activities of multiple players, where a single group activity is a function of the action of each and every player within the group~\cite{direkoglu2012team}. The activity of group can be observed as spontaneous emergent action, conducted by the activities and interactions of individuals within it. While the latter is the recognition of separate actions of multiple players in parallel, where two or more players participates. Figure~\ref{fig:def_GAR} shows the differences among individual action recognition (IAR), GAR, and MAR respectively. The GAR example (yellow box) shows that where without knowledge of all of the players in the opposite of the net, it is improbable that the algorithm will infer the accurate actions (e.g., if one of the player does not participate the blocking, the activity is ``double-block'' indeed). Only observing all subjects provides enough evidence for the correct recognition. Therefore, GAR is more challenging than individual action recognition, requiring to combine multiple computer vision techniques, such as player detection, pose estimation and ball tracking. Fig. \ref{fig:gar_framework} presents a typical framework for GAR.

An early work on group activity recognition is proposed by W. Choi \etal in 2009 \cite{choi2009they}. The proposed framework is composed of people detection, tracking, pose estimation, spatial-temporal local descriptor and classifier, where hand-crafted features -- HOG is adopted. Though W. Choi \etal only test the proposed framework on their own dataset for GAR, it inspires the following approaches.

A. Maksai \etal \cite{maksai2016players} propose a approach to model the interaction between players and ball for GAR. The proposed approach employs graphical models to track the ball and detect players, resulting in a player graph and a ball graph. In the player graph, each node represents a play location. With massage passing over the two graphs, the proposed approach can model the interaction between the ball and players. However, the main purpose of this work is ball tracking and the settings of GAR lack challenge, for example, there are only 4 classes of the ball state -- flying, passed, possessed and out of play.

In contrast, M. Ibrahim \etal \cite{ibrahim2016hierarchical} proposed a hierarchical deep model for GAR, where each player is detected first and the dynamics of each player are modeled using a LSTM, finally, the a group-level LSTM is adopted to aggregate all players' dynamics and makes a prediction. The hierarchical deep model achieves 51.1\% on \textbf{HierVolleyball} dataset and 81.9\% on \textbf{HierVolleyball-v2}.

T. Shu \etal \cite{shu2017cern} use a graph to model group activities, proposing a \emph{Confidence-Energy Recurrent Network} (CERN). Specifically, CERN first employs a tracker to obtain the trajectories of players and then constructs a graph, where each node represents an individual player position in a video frame and each edge represents the relationship between two nodes. Two types of LSTMs are applied -- node LSTM and edge LSTM to compute deep features of graph nodes and edges. CERN achieves 83.6\% on \textbf{HierVolleyball-v2}.

In contrast, T. Bagautdinov \etal \cite{bagautdinov2017social} proposed an end-to-end approach for GAR, where player detection and action recognition adopts a shared fully-connected CNN. The detection branch applies \emph{Markov Random Field} (MRF) to refine the predicted player positions and the classification branch uses a matching \emph{Recurrent Neural Network} (RNN) to predict individual's action and their group activity. Without extra tracking models, the proposed model takes less time for training and inference. In terms of the performance, it obtains 87.1\% accuracy on \textbf{HierVolleyball-v2}.

\begin{table}[t]
\centering
\caption{Deep learning model for group activity recognition in sports.}\label{tab:gar_models}
\scalebox{1}{
\begin{tabular}{|c|c|c|}
\hline
     Model & Venue &HierVolleyball-v2 \\ 
     \hline\hline
     M. Ibrahim \etal \cite{ibrahim2016hierarchical} &CVPR-2016 & 81.9 \\
     CERN~\cite{shu2017cern} &CVPR-2017 & 83.6 \\
     T. Bagautdinov \etal \cite{bagautdinov2017social} &CVPR-2017 &87.1\\
     RCRG \cite{ibrahim2018hierarchical} &ECCV-2018 &89.5 \\
     StageNet \cite{qi2019stagnet} &TCSVT &89.3\\ 
     POGARS \cite{thilakarathne2021pose} &Arxiv &93.9\\
     Anchor-Transformer \cite{gavrilyuk2020actor} &CVPR-2020  &94.4\\
     DIN~\cite{yuan2021spatio} &CVPR-2021 &93.1 \\
     Pose3D \cite{duan2021revisiting} &CVPR-2022 & 91.3\\
     \hline 
\end{tabular}
}
\end{table}

Similarly, RCRG \cite{ibrahim2018hierarchical} extend the two-stage framework in \cite{bagautdinov2017social,shu2017cern} via introducing a hierarchical relational network, which is similar to graph neural networks, \ie, the new representation of a node is obtained by aggregating the information of its neighbors. 

StageNet \cite{qi2019stagnet} is composed of 4 stages: player detection, semantic graph construction, temporal information integration and spatial-temporal attention. Player detection and semantic graph construction are similar to RCRG \cite{ibrahim2018hierarchical}, \ie, each node of the graph represent a player position and the edges represent the relationships determined by the spatial distance and temporal correlations among players. In terms of temporal information integration, structural RNNs -- node RNN and edge RNN are applied and finally the aggregated information is fed into spatial-temporal module. Using spatial-temporal attention makes StageNet more explainable.

Recently, the poses of players are introduced into GAR. H. Thilakarathne \etal \cite{thilakarathne2021pose} propose a \emph{Pose Only Group Activity Recognition System} (POGARS), which consists of two key modules -- player tracking and pose estimation and each player is represented by 16 2D keypoints. After that, POGARS stacks multiple temporal and spatial convolutioanl layers to obtain high-level player representations. In addition, POGARS investigates different person-level fusion approaches, including early fusion and late fusion. Finally, POGARS achieves 93.2\% accuracy on \textbf{HierVolleyball-v2} and the performance can be further improved to 93.9\% by using both player poses and ball tracklets. While Pose3D \cite{duan2021revisiting} adopts skeleton heatmaps instead of the 2D coordinates and the feature extraction model is a 3D CNN, achieving 91.3\% accuracy.

H. Yuan \etal \cite{yuan2021spatio} introduces dynamic relation (DR) and dynamic walk (DW) into GAR models, proposing a \emph{Dynamic Inference Network} (DIN), where the detected players are constructed into a spatial-temporal graph and then DR is used to predict the relationships among players and DW is used to predict the dynamic walk offset to allow global interaction over the entire spatial-temporal graph. Using DR and DW, DIN obtains 93.1\% on \textbf{HierVolleyball-v2}.

K. Gavrilyuk \etal \cite{gavrilyuk2020actor} propose a transformer based model -- Anchor-Transformer, where the representations of different players are fused via a transformer instead of a LSTM. Similarly, Anchor-Transformer first employs a player detection model to obtain the individuals and then fuses the individual embeddings using a transformer for classification. It achieves 94.4\% on \textbf{HierVolleyball-v2} using both pose and optical flow.

Apart from volleyball, GAR in football is also investigated. T. Tsunoda \etal \cite{tsunoda2017football} propose a hierarchical LSTM model to recognition football team activities, which is similar to the model in \cite{ibrahim2016hierarchical}, but the videos in football dataset is captured by multiple synchronized cameras.

Also, we present the performances of different models in Table \ref{tab:gar_models}. Note that most models conduct experiments on \textbf{HierVolleyball-v2}, thus, we only report the performance on this dataset. And the proposed models are flexible and can be transferred into other team sports like football and basketball.

\section{Applications}\label{sec:application}
As aforementioned, the video action recognition in sports spawn a wide sorts of applications in our daily life. We categorize the applications into the following aspects,
\begin{itemize}
    \item \textbf{Training Aids}: Since the sports video corpus contains a large amount of historical records of competition and training clips, it is a good source of information for sports coaches and players to analyze and extract useful tactics. As one of the most common approaches, the video action recognition can provide a straightforward way to obtain the actions/events (i.e., the basic unit of sports). Then, the actions sequences/combinations could be correlated with the wining strategies, which can either guide the training of players or help with designing the game plan. For example,~\cite{fani2017hockey} introduces an action recognition hourglass network (ARHN) to interpret players actions in ice hockey video, where the recognized hockey players' poses and hockey actions are valuable pieces of information that potentially can help coaches in assessing player's performance. Another well-known case for training aid is sports AI coach system~\cite{wang2019ai}, which can provide personalized athletic training experiences based on the video sequences. The action recognition is one of the key step in AI coach system to support complex visual information extraction and summarization.    
    
    \item \textbf{Game Assistance (Video Judge)}: The video-based game judge has been widely involved in the modern sports video analysis systems, where most of the system adopt the action recognition as the elementary module.~\cite{nekoui2020falcons} proposes a virtual referring network to evaluate the execution of a diving performance. This assessment is based on visual clues as well as the body actions in sequences. Upon the same sports (diving),~\cite{parmar2019and} comes up with a idea to learn spatio-temporal features that explain the related tasks such as fine-grained action recognition, so as to improve the action quality assessment. Rather than judge the performance of the athlete via the action recognition,~\cite{pan2020hierarchical} develops a sports referee training system, which intends to recognize whether a trainee makes the proper judging signals. In this work, a deep belief network is adopted to capture high quality features for hand gesture recognition. 
    
    \item \textbf{Video Highlights}: Highlights segmentation and summarization in sports videos are with a wide viewership and has a great amount of commercial potential. While the foundation for accomplishing this goal is the action recognition step in processing the sports video. As a typical example,~\cite{nakano2020estimating} proposes an automatic highlight detection method to recognize the spatio-temporal pose in skating videos. Through an accurate action recognition module, the proposed method is capable of locating and stitching the target figure skating poses. Since the jumps in figure skating sports are one of the most eye-catching actions/poses, it appears commonly in the highlight clips of figure skating sports, where~\cite{tian2020multi} dedicates to recognizing the 3D jump actions and recovering the poor-visualising actions. Another work~\cite{shroff2010video} treats the video highlights as a combinatorial optimization problem, and regards the diversity of recognized action as one of the constrains. To maximize the diversity and lower the recognition error, the overall quality of the highlights video is improved drastically.  
    
    \item \textbf{Automatic Sports News Generation (ASNG)}: There is a large demand of sports news generation. Existing ASNG systems normally adopt the statistical numbers in matches, such as the number of shots, corners and free kicks in a football match and then use texts to describe the numbers \cite{kanerva2019template,gong2017automatic}. However, in many cases, the numbers are provided by human instead of automatically recognized in videos, which is time-consuming and a massive workload. While video action recognition techniques can automatically generate these numbers and only require a few people to verify the final results, saving time and reducing workload. Plus, thanks to the technique of visual captioning, \ie, using texts to describe images \cite{wang2020neighbours,wang2020diversity,wang2022distinctive} and videos \cite{chen2021motion,zhang2021open}, we can also directly generate textural descriptions from videos. Nevertheless, recognizing the actions of players is still required, since better recognition results can significantly improve the naturalness, fluency and accuracy of the final texts.
    
    \item \textbf{General Research Purposes}: As one of the main branches of video analysis, the action recognition is never stopped being studied. We can observe that the sports videos account for a significant portion of the target video categories~\cite{ramanathan2014human,herath2017going,pareek2021survey,kong2022human,wu2017recent}. Not surprisingly, the sports video analysis has been a very popular research topic, due to the variety of application areas, ranging from analysis of athletes’ performances and rehabilitation to multimedia intelligent devices with user-tailored digests. Datasets (videos)~\cite{pers2005cvbase,rodriguez2008action,de2008distributed,parisot2019consensus,d2009semi,li2010action,niebles2010modeling,bermejo2011violence,de2011evaluation,de2011evaluation,gourgari2013thetis} focused on sports activities or datasets including a large amount of sports activity classes are now available and many research contributions benchmark on those datasets. A large amount of work is also devoted to fine-grained action recognition through the analysis of sports gestures/poses using motion capture systems. On the other hand, the ability to analyze the actions which occur in a video is essential for automatic understanding of sports. The action recognition techniques can efficiently collect and classify the actions/events in sports video, and consequently help a lot with the sports statistics analysis which is the basis to understand the sports~\cite{meng2022analysis,li2022video,carson2008utilizing,shih2017survey,soomro2014action,liu2014research}. 
\end{itemize}

All in all, the application of the video action recognition in sports are widely spread in different purposes and draws more attention from either sports domains or computer vision domains. In the next section, we will go through the possible challenges when applying the action recognition in realistic sports videos.

\section{Challenges}\label{sec:challenge}
In this section, we summarize the challenges when applying those action recognition baselines on sports videos in practical. Specifically, we categorize the challenges into the following aspects,
\begin{itemize}
    \item \textbf{Data Collection and Annotation}. As one of the crucial step for establishing a dataset for further research, data collection and annotation draw more attention and the qualities of them directly affect the performance of the action recognition task~\cite{zhang2016action,zhang2016rgb,carreira2017quo}. However, the main difference of sports datasets comparing to other human action recognition datasets (e.g., ActivityNet, Kinetics400, and UCF101) in terms of collections and annotations are 1) Accessibility: Most of the representative sports videos comes from the untrimmed live broadcasting clips, which is access-restricted due to the authorship or the copyright of the clips. While the self-recorded sports videos are with comparably lower quality either in footage resolution (without best angle) or the content itself (e.g., the target players are amateurish), such datasets can lead to the inefficient training of the action recognition algorithms, which generates models with poor generalization ability in practical task; 2) Expertise: Since the sports videos normally focus on specific sports category (e.g., hockey, volleyball, and figure skating), the annotation requires a higher expertise than the regular human actions (e.g., walk, run, and sit). The more professional the annotators are especially in the target sports domain, the better the quality of the annotations are, which leads to promising performance of action recognition algorithms in real inference tasks. One possible direction is using active learning approaches \cite{zhan2022comparative,zhan2021multiple,zhancomparative} to reduce the workload of annotation; 3) Multi-purpose: As a general trend, the video dataset for the actions recognition is rarely with only one purpose, so are sports datasets. Some of the video datasets~\cite{chen2019relation,megrhi2016spatio} also are designed to accomplish the temporal action localization, spatio-temporal action localization, and complex event understanding. To serve multiple purposes, the author of the dataset needs to prepare a variety of labeling content and auxiliary feature information, which is even challenging for sports videos due to the specific nature of the actions. For example, extracting the skeleton feature from table tennis video is difficult due to the dense and fast-moving nature of the stroke actions. Compared to the general human actions recognition datasets, the sports action recognition datasets usually take more efforts to be established and developed.

    \item \textbf{Dense and Fast-moving Actions}. One the one hand, the traditional action recognition baselines~\cite{gan2016you,wang2018temporal,long2018attention,lin2019tsm,bertasius2021space} are designed to tackle those actions around $4\sim20$ (or over 20s as an event) seconds on average, where some of the actions in sports video are out of this range. For example, the stroke action in the table tennis competition commonly task only $0.4\sim 2$ seconds via a conventional broadcasting camera. Fast-moving characteristics requires the action recognition algorithms to capture a relatively short-lived events from the video stream and tolerate with the background changes which is easy to confuse the judgement in such scenario~\cite{hao2013human,anuradha2016spatio}. On the other hand, as the nature of the table tennis sports itself, two players takes action to stoke the ball in turns until one of the player wins a point, where the stoke actions are in a super dense distribution compared to other sports (e.g., soccer and basketball). There could be 8 to 10 stroke actions in less than 6 seconds, which means the action recognition algorithms should be more sensitive to the boundary of two actions and it is proved to be a challenging task for some of the state of the art models~\cite{karpathy2014large,yue2015beyond,donahue2015long,srivastava2015unsupervised}. Although, we can fine-tune the baselines carefully on the video datasets with dense and fast-moving actions, the performance is still far less than expectation~\cite{lorre2020temporal,ghadiyaram2019large} compared to those regular action recognition tasks. Thus, the sports with fast-moving and dense actions are potential to be further explored in action recognition domain and could be a basis for developing more robust recognition algorithms. 
    
    \item \textbf{Camera Motion, Cut and Occlusion.} The main difference of video datasets and still image datasets are the motion of target object, where the quality of the motion features may affect the action recognition performance~\cite{wang2014action,lee2018motion,fathi2008action}. The traditional way to form motion trajectories heavily relies on the extraction of optical flow~\cite{sevilla2018integration,piergiovanni2019representation}, where most of them are based on the video recorded by fixed camera with the complete and clear view of objects. However, in recent sports videos/streaming, the camera motion is no longer fixed and tend to be variant since the highlights of the video keeps changing (e.g., the zoom-in and zoom-out highlights). This naturally leads to the cut of view and more or less occlusion in the recorded videos/streaming, which causes challenges to those well-established action recognition benchmarks~\cite{simonyan2014two,feichtenhofer2016convolutional,wang2018temporal,tran2018closer,zhou2018temporal,lin2019tsm,liu2020fsd,wang2020makes} (e.g., those algorithms are barely tolerable to the data sample from different camera motions, with cut and occluded objects). Although there exists work~\cite{wang2013action,jain2013better,wang2013dense} to take the camera motion into considerations when designing the motion descriptor for action recognition task, the cut and occluded objects are still a problem which makes the feature space inconsistent. Several works~\cite{weinland2010making,angelini20192d,iosifidis2012multi} intend to solve the occlusion problem individually by modifying the structure and attention of the motion descriptor, where it is limited to single target and we know that sports videos commonly involve multiple players, which increases the complexity when applying these occlusion-handling methods.   
    
    \item \textbf{Long-tailed Distribution and Imbalanced Data.} Before applying action recognition algorithms on the video datasets, we normally check the statistics of the dataset in case of any undesirable situation such as the long-tailed distribution of the target actions. As we know, the long-tailed learning~\cite{ouyang2016factors,zhang2017range,zhang2021deep} is one of the most challenging problems in visual recognition, aims to train well-performing models from a large number of frames that follow a long-tailed class distribution. Unfortunately, sports datasets such as soccer, basketball, and table tennis suffer a lot from such long-tailed class distribution and imbalance, which degrades the model performance drastically~\cite{zhang2021videolt,sozykin2018multi,ding2017facial,wu2016mixed}. This common status quo in sports video datasets motivate us to either adopt a proper data augmentation method prior training or design a robust action recognition algorithm to mitigate the negative effects of long-tailed distribution. As shown in Figure~\ref{fig:long-tailed}, we briefly compare the distribution of classes in general video recognition versus the distribution in long-tailed video recognition. Further we showcase two representative datasets, which are table tennis videos (P$^2$A~\cite{p2a2022} dataset) and the sports video in wild (SVW~\cite{p2a2022} dataset). The middle and bottom figure demonstrate the class of action in untrimmed sports video commonly follow a long-tailed distribution and naturally form imbalanced datasets.
    
\begin{figure}[t]
\centering
\includegraphics[width=0.8\linewidth]{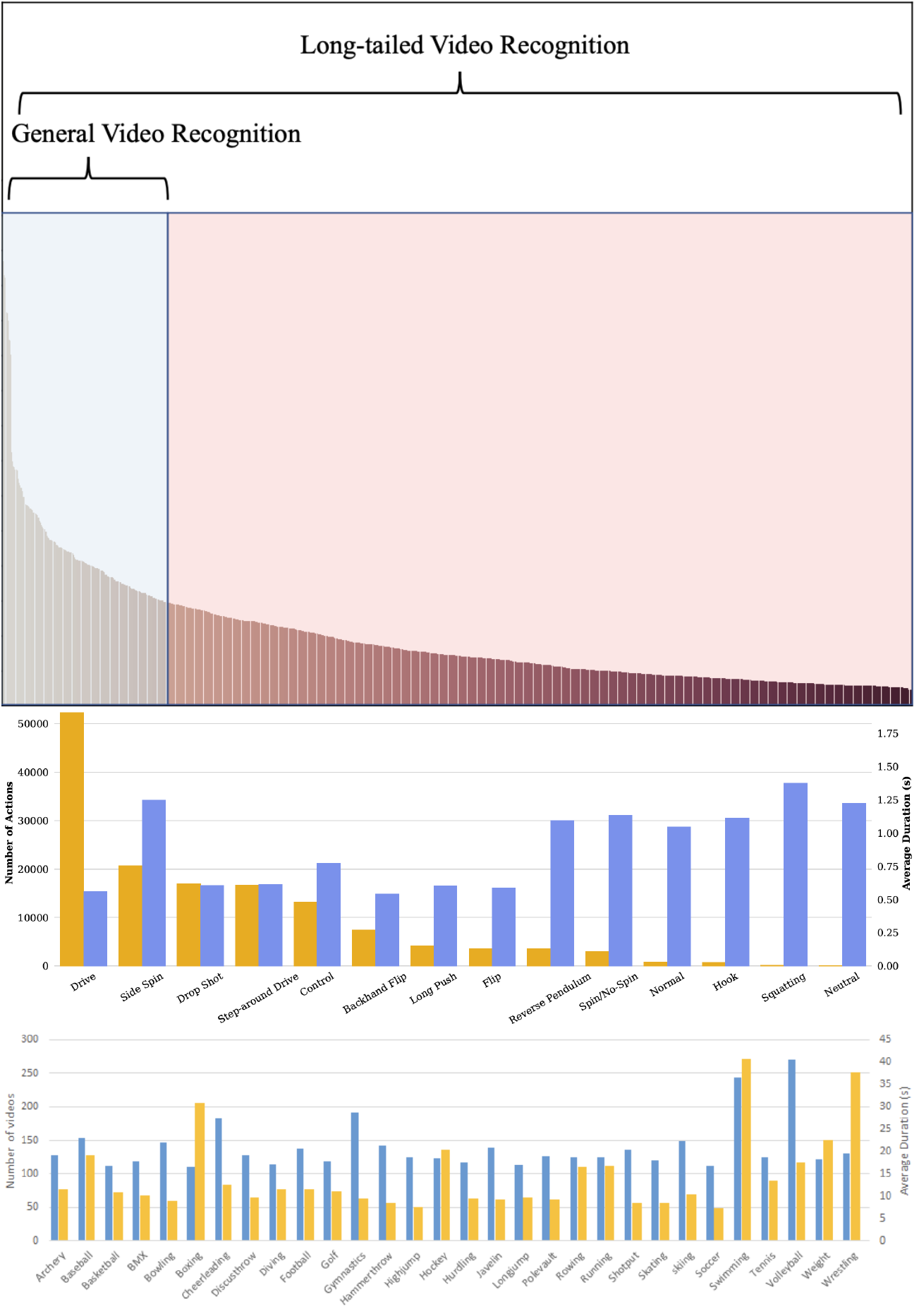}
\caption{Example of Long-tailed Distribution and Imbalanced Data. Top: Long-tailed vs General~\cite{zhang2021videolt}; Middle: The long-tailed distribution of classes in P$^2$A dataset~\cite{p2a2022}; Bottom: The imbalanced classes in SVW dataset~\cite{safdarnejad2015sports}.}
\label{fig:long-tailed}
\end{figure}
    
    \item \textbf{Multi-camera and Multi-view Action Recognition.} As we mentioned in Applications section, the action recognition techniques are widely used in web or TV streaming for the purpose of Video Highlights. While the videos are normally recorded via multiple cameras and are in different views~\cite{karpathy2014large,shao2020finegym,p2a2022,liu2020fsd,li2018resound,faulkner2017tenniset,rodriguez2008action}, this requires the robustness and adaptability of the corresponding action recognition algorithms. Via a thorough investigation in this paper, most of the benchmarks~\cite{karpathy2014large,yue2015beyond,donahue2015long,srivastava2015unsupervised,gan2016you,wang2018temporal,long2018attention,lin2019tsm,bertasius2021space} of action recognition on video datasets focus on single-camera or single-view actions, where it does not conform with the format of sports videos. Although some of the action recognition algorithms~\cite{wang2019generative,wang2018dividing,hao2017multi} intend to split the task into several sub-tasks (i.e., training separately on each view) and combine the results for a performance promotion, it is still challenging to detect and switch the sub-models between each view when handling a complete sports video.

    \item \textbf{Transfer, Few-shot and Zero-shot Learning.} To ensure the accuracy of action recognition, there frequently needs to collect a large number of video clips, extract frames from clips, and annotate frames with fine-grained labels (such as temporal labels and/or skeletons). The data collection and annotation thus becomes extremely expensive, when sports of multiple categories are desired. Yet another way to lower the cost of action recognition from sport videos is to pre-train backbone models using videos collected from a wide spectrum of sport categories in a self-supervised manner~\cite{tung2017self,sermanet2018time,hu2021contrast} and then fine-tune~\cite{li2018delta,wan2019towards,li2020rifle,xiong2022grod} the pre-trained model using few labeled samples for the target sport analytic tasks, so as to transfer the knowledge of video understanding to specific sport action recognition tasks. Thus, few-shot and even zero-shot learning~\cite{xu2015semantic,mishra2018generative,bo2020few,zhang2020few} are requested to generalize action recognition tasks by incorporating labeled samples and/or explicit domain knowledge~\cite{wang2019survey}.
\end{itemize}

\section{Conclusion}\label{sec:conclusion}
In this paper, we review and survey the works on video action recognition for sports analytics. We cover dozens of sports, categorized into two streams \emph{(1) team sports}, such as football, basketball, volleyball, hockey and \emph{(2) individual sports}, such as figure skating, gymnastics, table tennis, tennis, diving and badminton. Specifically, we present numerous existing solutions, such as statistical learning-based methods for traditional computer visions, deep learning-based methods with 2D and 3D neural models, and skeleton-based methods using auxiliary information, all for sports video analytics. We compare the performance of these methods using literature reviews and experiments, where we clearly illustrate the status quo on performance of video action recognition for both team sports and individual sports. Finally, we discuss the open issues, including technical challenges and interesting problems, in this area and conclude the survey. To facilitate the research in this field, we release a toolbox for sport video analytics for public research.


\bibliographystyle{IEEEtran}
\bibliography{myref}

\end{document}